\documentclass[10pt,twocolumn,letterpaper]{article}
\usepackage{subfiles}

\usepackage{xr-hyper}       
\usepackage{xr}
\makeatletter
\newcommand*{\addFileDependency}[1]{
  \typeout{(#1)}
  \@addtofilelist{#1}
  \IfFileExists{#1}{}{\typeout{No file #1.}}
}
\makeatother

\newcommand*{\myexternaldocument}[1]{%
    \externaldocument{#1}%
    \addFileDependency{#1.tex}%
    \addFileDependency{#1.aux}%
}
\myexternaldocument{appendix_1067}

\usepackage[pagenumbers]{cvpr} 

\usepackage{graphicx}
\usepackage{amsmath}
\usepackage{amssymb}
\usepackage{booktabs}
\usepackage{multirow, makecell} 
\usepackage{bbding}
\usepackage{float}
\usepackage{url}

\usepackage[pagebackref,breaklinks,colorlinks]{hyperref}

\usepackage[capitalize]{cleveref}
\crefname{section}{Sec.}{Secs.}
\Crefname{section}{Section}{Sections}
\Crefname{table}{Table}{Tables}
\crefname{table}{Tab.}{Tabs.}

\def\cvprPaperID{1067} 
\def\confName{CVPR}
\def\confYear{2022}

\usepackage{amsmath,amsfonts,bm}

\def\eqref#1{equation~\ref{#1}}

\def\1{\bm{1}}

\def\vb{{\bm{b}}}
\def\vc{{\bm{c}}}
\def\vd{{\bm{d}}}

\def\vo{{\bm{o}}}

\def\vv{{\bm{v}}}
\def\vw{{\bm{w}}}
\def\vx{{\bm{x}}}

\def\vz{{\bm{z}}}

\def\mS{{\bm{S}}}

\def\mW{{\bm{W}}}
\def\mX{{\bm{X}}}
\def\mY{{\bm{Y}}}

\DeclareMathAlphabet{\mathsfit}{\encodingdefault}{\sfdefault}{m}{sl}
\SetMathAlphabet{\mathsfit}{bold}{\encodingdefault}{\sfdefault}{bx}{n}

\newcommand{\R}{\mathbb{R}}

\begin{document}

\title{CIPS-3D: A 3D-Aware Generator of GANs\\Based on Conditionally-Independent Pixel Synthesis}

\author{%
Peng Zhou$^1$, Lingxi Xie$^2$, Bingbing Ni$^{1}$, Qi Tian$^2$\\
$^1$Shanghai Jiao Tong University,\quad$^2$Huawei Inc.\\
{\tt\small \{zhoupengcv, nibingbing\}@sjtu.edu.cn}, {\tt\small 198808xc@gmail.com}, \\  {\tt\small tian.qi1@huawei.com}
}



\twocolumn[{%
      \maketitle
      \vspace{-1.3cm}
      \centering
      \begin{figure}[H]
        \hsize=\textwidth
        \footnotesize
        \centering
        \renewcommand{\tabcolsep}{2pt} \renewcommand{\arraystretch}{2}
        \resizebox{\textwidth}{!}{%
          \begin{tabular}[t]{l|r}%
            \resizebox{!}{6cm}{%
              \begin{tabular}[]{c}
                \includegraphics[width=\linewidth]{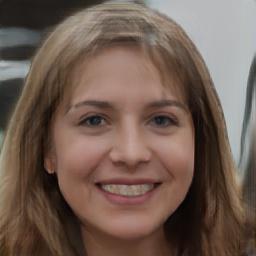}
                \includegraphics[width=\linewidth]{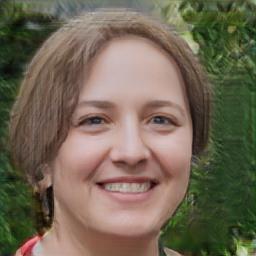}
                \\
                \resizebox{!}{0.7cm}{(a) GIRAFFE~\cite{niemeyer2021GIRAFFE}}
                \\
                \includegraphics[width=\linewidth]{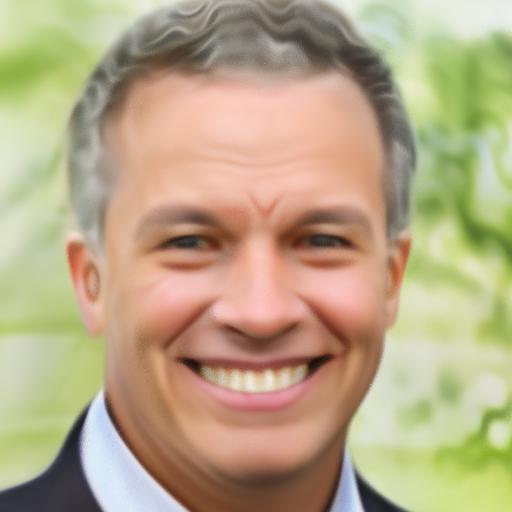}
                \includegraphics[width=\linewidth]{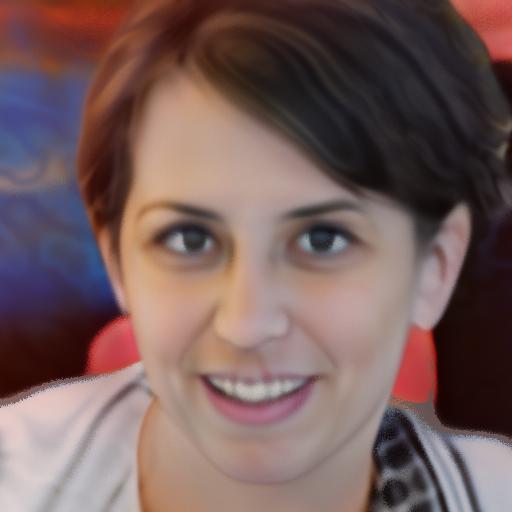}
                \\
                \resizebox{!}{0.7cm}{(b) pi-GAN~\cite{chan2021piGAN}}
              \end{tabular}%
            } \hspace{0.02cm} & \hspace{0.02cm}
            \resizebox{!}{6cm}{%
              \centering
              \renewcommand{\tabcolsep}{5pt} \renewcommand{\arraystretch}{2}
              \begin{tabular}[]{ccc|c|cc}
                \includegraphics[width=\linewidth]{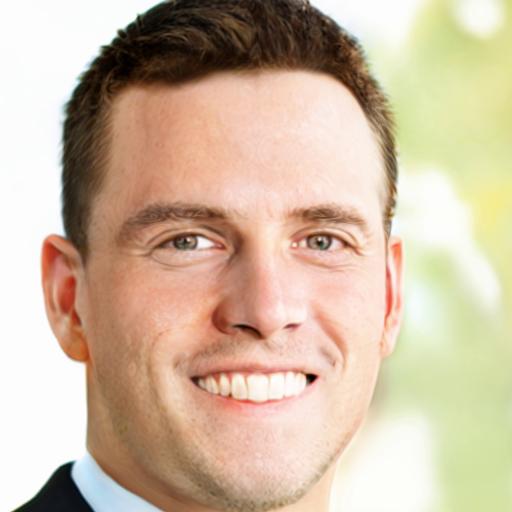}    &
                \includegraphics[width=\linewidth]{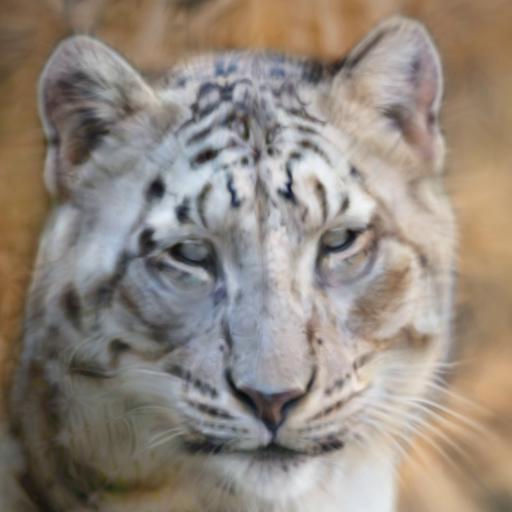}    &
                \includegraphics[width=\linewidth]{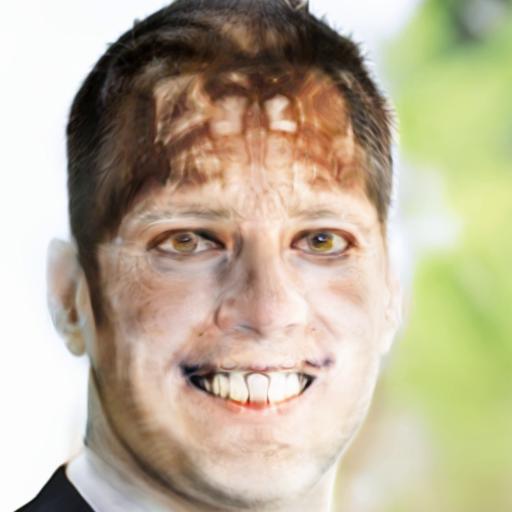}                      &
                \includegraphics[width=\linewidth]{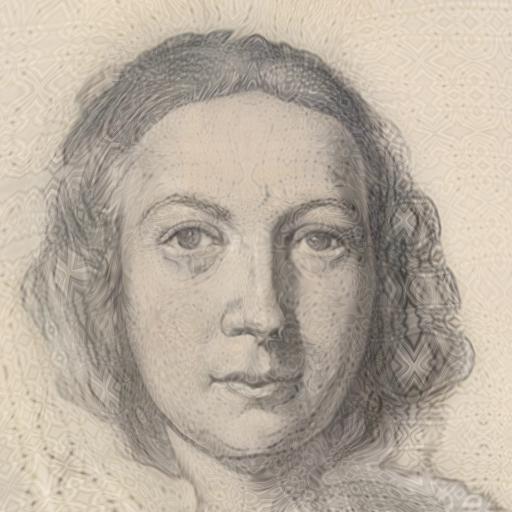}                      &
                \includegraphics[width=\linewidth]{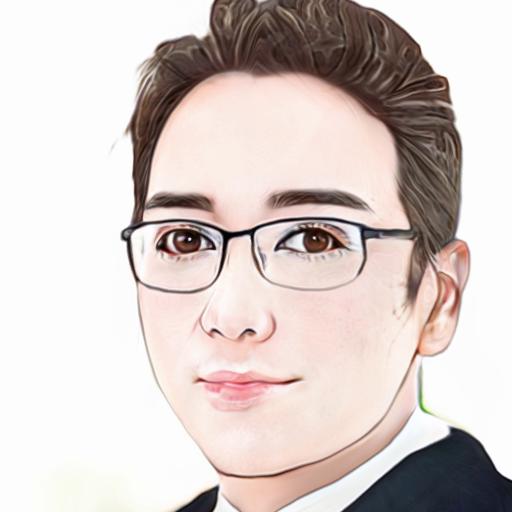}                 &
                \includegraphics[width=\linewidth]{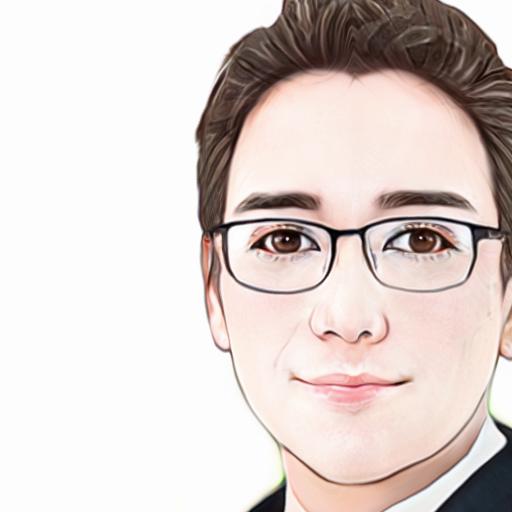}
                \\ 
                \includegraphics[width=\linewidth]{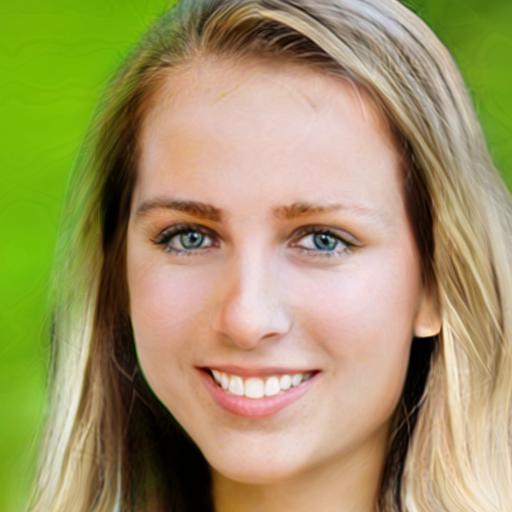}                      &
                \includegraphics[width=\linewidth]{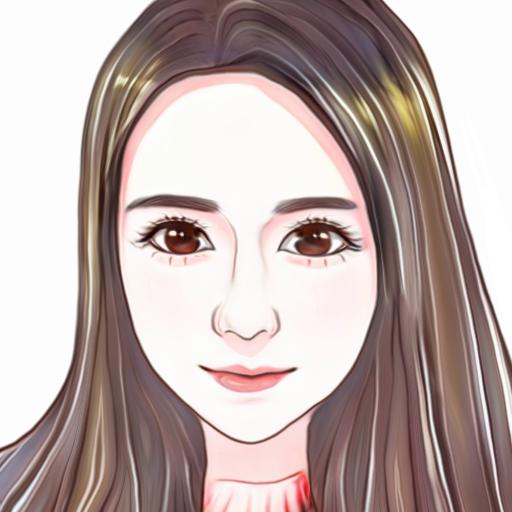}                      &
                \includegraphics[width=\linewidth]{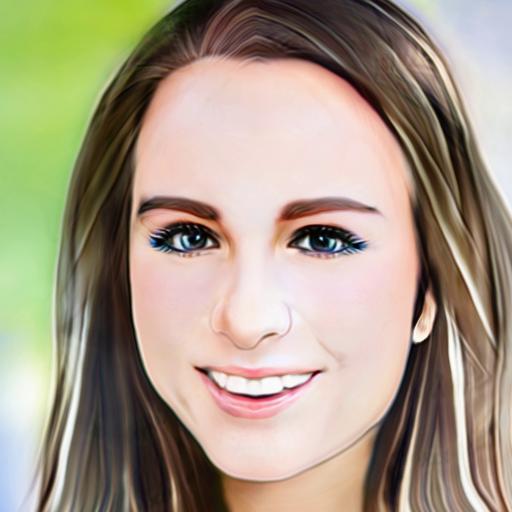} &
                \includegraphics[width=\linewidth]{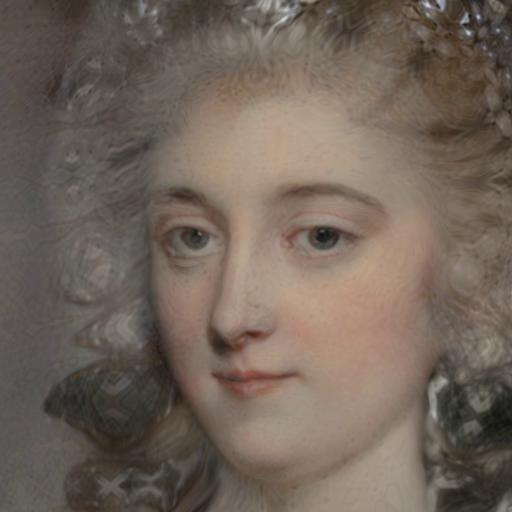}                      &
                \includegraphics[width=\linewidth]{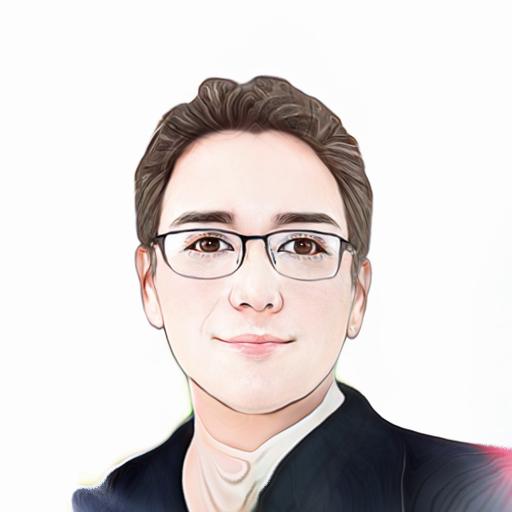}                 &
                \includegraphics[width=\linewidth]{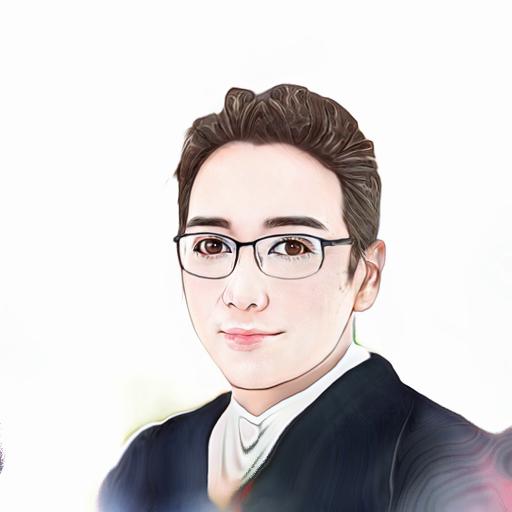}
                \\
                \resizebox{!}{0.7cm}{Content}                                                                                & \resizebox{!}{0.7cm}{Style} & \resizebox{!}{0.7cm}{Stylization} & \resizebox{!}{0.7cm}{MetFaces} & \multicolumn{2}{c}{\resizebox{!}{0.7cm}{Stylization and multiple views}}
                \\
                \\
                \multicolumn{6}{c}{              \resizebox{!}{0.7cm}{(c) All these images are synthesized by CIPS-3D at the resolution of $512\times 512$.}}
              \end{tabular}%
            }
          \end{tabular}%
        }
        \caption{Three types of 3D-aware GANs. (a): There are apparent artifacts in the images generated by GIRAFFE~\cite{niemeyer2021GIRAFFE}. (b): The images generated by pi-GAN~\cite{chan2021piGAN} are blurred and lack details. (c): CIPS-3D can generate photo-realistic high-fidelity images. We fine-tune the base model trained on FFHQ so that the transferred model can generate other types of style images. Then we interpolate the base model and the transferred model to create a new model that can generate stylized images. CIPS-3D enables one to manipulate the pose of the stylized faces (the rightmost images) explicitly. For details, please refer to~\cref{sec:finetune,sec:interpolation}.}
        \label{fig:first_page_fig}
      \end{figure}
      \vspace{-0.3cm}
    }]

\begin{abstract}
  \vspace{-0.3cm}
  The style-based GAN (StyleGAN) architecture achieved state-of-the-art results for generating high-quality images, but it lacks explicit and precise control over camera poses. The recently proposed NeRF-based GANs made great progress towards 3D-aware generators, but they are unable to generate high-quality images yet. This paper presents \textbf{CIPS-3D}, a style-based, 3D-aware generator that is composed of a shallow NeRF network and a deep implicit neural representation (INR) network. The generator synthesizes each pixel value independently without any spatial convolution or upsampling operation. In addition, we diagnose the problem of mirror symmetry that implies a suboptimal solution and solve it by introducing an auxiliary discriminator. Trained on raw, single-view images, CIPS-3D sets new records for 3D-aware image synthesis with an impressive FID of $6.97$ for images at the $256\times256$ resolution on FFHQ. We also demonstrate several interesting directions for CIPS-3D such as transfer learning and 3D-aware face stylization. The synthesis results are best viewed as videos, so we recommend the readers to check our github project at \url{https://github.com/PeterouZh/CIPS-3D}.

\end{abstract}

\vspace{-0.3cm}
\section{Introduction}
\label{sec:intro}

Generative Adversarial Networks (GANs)~\cite{goodfellow2014Generative} can synthesize high-fidelity images~\cite{brock2018Large,karras2017Progressive,karras2019StyleBased,karras2019Analyzing,karras2020Training,karras2021AliasFree} but lack an explicit control mechanism to adjust the viewpoint for the generated object. Previous methods alleviate this problem by finding the latent semantic vectors existing in pre-trained 2D GAN models~\cite{jahanian2020steerabilitya,shen2020Interpreting,harkonen2020GANSpace,shen2020ClosedForm}. However, these methods can only roughly change the pose of the object implicitly and fail to render the object from arbitrary camera poses. Several 3D-aware methods have been proposed to enable explicit control over the camera pose~\cite{zhu2018Visual,henzler2019Escaping,nguyen-phuoc2019HoloGAN}. However, these models are often limited to low-resolution images with disappointing artifacts.

Recently, there has been a growing interest in leveraging the neural radiance fields (NeRF)~\cite{mildenhall2020NeRF} to build 3D-aware GANs. To our knowledge, there exist two types of 3D-aware generators: (i) using a pure NeRF network as the generator~\cite{schwarz2020GRAF,chan2021piGAN}; (ii) generating low-resolution feature maps with a NeRF network and then upsampling with a 2D CNN decoder~\cite{niemeyer2021GIRAFFE,anonymous2021StyleNeRF}. The former suffers from a low capacity of the generator because the NeRF network is memory-intensive, which limits the depth of the generator. Thus the synthesized images are blurred and lack sharp details (see~\cref{fig:first_page_fig}b). The latter is susceptible to aliasing due to non-ideal upsampling filters (see~\cref{fig:first_page_fig}a)~\cite{karras2021AliasFree,parmar2021Buggy}.

This paper presents CIPS-3D, an approach to synthesize each pixel value independently, just as its 2D version did~\cite{anokhin2021Image}. The  generator consists of a shallow 3D NeRF network (to alleviate memory complexity) and a deep 2D implicit neural representation (INR) network (to enlarge the capacity of the generator)~\cite{park2019DeepSDF,mescheder2019Occupancy,chen2019Learninga}, without any spatial convolution or up-sampling operations. Interestingly, the design of our generator is consistent with the well-known semantic hierarchical principle of GANs~\cite{bau2018GAN,yang2020Semantic}, where the early layers (\ie, the shallow NeRF network in our generator) determine the pose, and the middle and higher layers (\ie, the INR network in our generator) control semantic attributes and color scheme, respectively. The early NeRF network enables us to control the camera pose explicitly.

We found that CIPS-3D suffers from a mirror symmetry problem, which also exists in other 3D-aware GANs such as GIRAFFE~\cite{niemeyer2021GIRAFFE} and StyleNeRF~\cite{anonymous2021StyleNeRF}. Rather than simply attributing this issue to the dataset bias, we explain why this problem exists. Going one step further, we propose to utilize an auxiliary discriminator to regularize the output of the NeRF network, thus successfully solving this problem (see~\cref{sec:mirror_symmetry}). To train CIPS-3D at high resolution, we propose a training strategy named partial gradient backpropagation. Moreover, we provide a more efficient implementation for the Modulated Fully Connected layer (ModFC)~\cite{karras2019Analyzing,anokhin2021Image} to speed up the training (see ~\cref{sec:deep_inr}).

We validate the advantages of our approach on high-resolution face datasets including FFHQ~\cite{karras2019StyleBased}, MetFaces~\cite{karras2020Training}, BitmojiFaces~\cite{BitmojiFaces}, CartoonFaces~\cite{CartoonFaces}, and an animal dataset, AFHQ~\cite{choi2020StarGAN}. For 3D-aware image synthesis, CIPS-3D achieves state-of-the-art FID scores of $6.97$ and $12.26$ on FFHQ at $256^2$ and $1024^2$ resolution, respectively, surpassing the StyleNeRF~\cite{anonymous2021StyleNeRF} proposed very recently. Moreover, we verify that CIPS-3D works pretty well in transfer learning settings and show its application for 3D-aware face stylization (see~\cref{sec:finetune,sec:interpolation}). We will release the code to the public. We hope that CIPS-3D will serve as a good base model for downstream tasks such as 3D-aware GAN inversion and 3D-aware image editing.

\section{Related Work}

Implicit neural representation is a powerful tool for representing scenes in a continuous and memory-cheap way compared to mesh/voxel-based ones. It is usually implemented by a multilayer perception (MLP). The implicit representation has been widely applied in 3D tasks~\cite{chen2019Learninga,park2019DeepSDF,mescheder2019Occupancy,saito2019PIFu,littwin2019Deep,genova2019Learning,genova2020Local} as well as some 2D tasks such as image generation~\cite{skorokhodov2021Adversarial,anokhin2021Image} and super-resolution~\cite{chen2021Learning,xu2021UltraSR}. Equipped with volume rendering~\cite{kajiya1984Ray}, NeRF-based methods~\cite{mildenhall2020NeRF,martin-brualla2021NeRF,zhang2020NeRF,chen2021MVSNeRF,wang2021NeRF,barron2021MipNeRF,jang2021CodeNeRFa,yang2021Learning} enable novel view synthesis by learning an implicit function for a specific scene. Recently, there has been a trend combining NeRF with GANs~\cite{goodfellow2014Generative,arjovsky2017Wasserstein,gulrajani2017Improved,radford2015Unsupervised} to design 3D-aware generators~\cite{schwarz2020GRAF,chan2021piGAN,niemeyer2021GIRAFFE,anonymous2021StyleNeRF,niemeyer2021CAMPARI,devries2021Unconstraineda}.

Like GIRAFFE~\cite{niemeyer2021GIRAFFE} and StyleNeRF~\cite{anonymous2021StyleNeRF}, CIPS-3D utilizes NeRF to render features instead of RGB colors. However, our method differs from GIRAFFE and StyleNeRF in several ways. Both GIRAFFE and StyleNeRF adopt a two-stage strategy, where they render low-resolution feature maps first and then upsample the feature maps using a CNN decoder. CIPS-3D synthesizes each pixel independently without any up-sampling and spatial convolution operations. Moreover, CIPS-3D represents 3D shape and appearance with NeRF and INR networks, respectively, which is convenient for transfer learning settings.



\section{Method}

\begin{figure*}[!t]
  \begin{center}
    \includegraphics[width=\linewidth]{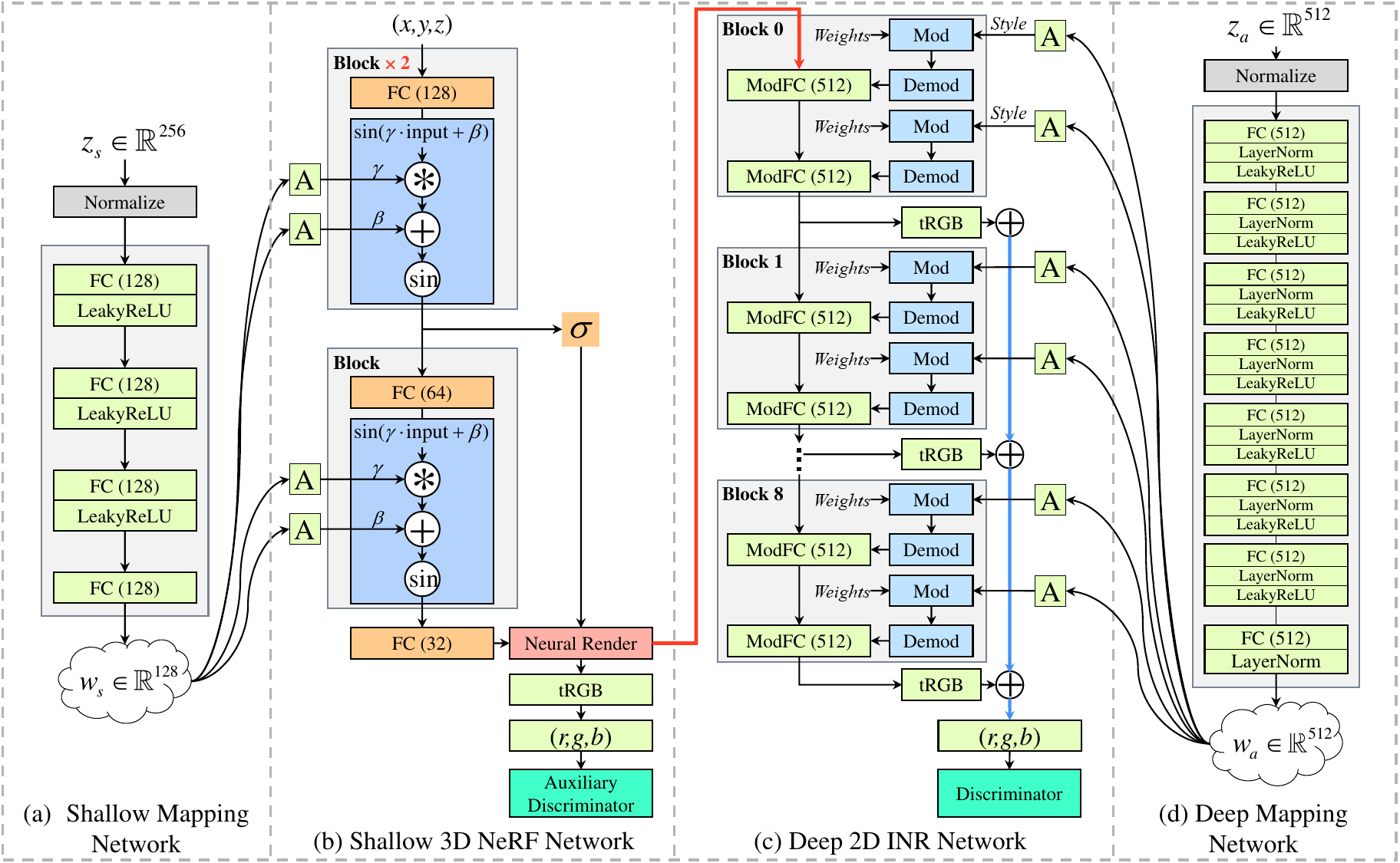}
  \end{center}
  \vspace{-0.65cm}
  \caption{The style-based 3D-aware generator with detailed hyperparameters. The NeRF network is shallow to save runtime memory. The INR network is deep to increase the capacity of the generator. We disentangle 3D shape and appearance, where the NeRF network is responsible for the 3D shape and the INR network for appearance. The auxiliary discriminator helps to overcome the problem of mirror symmetry (see~\cref{sec:mirror_symmetry}). For the INR network, each ModFC is followed by a LeakyReLU (not shown here). }
  \vspace{-0.5cm}
  \label{fig:framework}
\end{figure*}

\subsection{NeRF Network for 3D Shape}

\paragraph{Shallow NeRF Network}

A neural radiance field~\cite{mildenhall2020NeRF} is a continuous function $f$ whose inputs are 3D coordinates $\vx=(x, y, z)$ and viewing direction $\vd$, and whose outputs are emitted colors $\vc=(r, g, b)$ and volume density $\sigma$. A multi-layer perceptron (MLP) is usually used to parameterize the continuous function $f$:
\begin{equation}
  \resizebox{0.9\hsize}{!}{$
      f: \R^{\mathrm{dim}(\vx)} \times \R^{\mathrm{dim}(\vd)} \rightarrow \R^+ \times \R^3,\quad
      (\vx, \vd) \mapsto(\sigma, \vc).
    $}
\end{equation}
The 3D coordinates are sampled along camera rays. Each ray corresponds to a pixel of the rendered image. Thus to render a high-resolution image of size $H \times W$, the number of rays is large (\ie, $H \times W$). Besides, to obtain accurate 3D shape, many points need to be sampled on each ray. As a result, the NeRF network is memory-intensive.

To alleviate memory complexity, we adopt a shallow NeRF network to represent 3D shape while assigning the task of synthesizing high-fidelity appearance to a deep 2D INR network. In particular, a shallow NeRF network (see Fig.~\ref{fig:framework}b) containing only three SIREN blocks~\cite{sitzmann2020Implicita} is used as the initial layers of the GAN's generator.

\paragraph{Modulated SIREN Block}
The vanilla NeRF is restricted to a specific scene with fixed geometry. Because the generated images are various for GANs, the shape of each image is different. To render different shapes with one NeRF, we condition the NeRF network on a noise vector $\vz_s$ so that different shapes can be obtained by sampling  $\vz_s$. In particular, like StyleGAN~\cite{karras2019StyleBased}, we use a mapping network $m_s: \mathcal{Z}_s \rightarrow \mathcal{W}_s$ to map $\vz_s$ to $\vw_s$, and use $\vw_s$ to modulate the feature maps of the NeRF network (see~\cref{fig:framework}a).

We adopt the strategy of pi-GAN~\cite{chan2021piGAN}: modulating features with FiLM~\cite{perez2017FiLM,48757}, followed by a SIREN activation function~\cite{sitzmann2020Implicita}. The modulated SIREN block is given by
\begin{equation}
  \phi\left(\vx \right)=\sin \left(\gamma \cdot (\mW\vx+\vb) + \beta \right),
\end{equation}
where $\gamma=\mathrm{Affine}(\vw_s)$ and $\beta=\mathrm{Affine}(\vw_s)$ represent frequency and phase, respectively. The block contains a FC layer with $\mW$ and $\vb$ as the weight matrix and the bias. As shown in~\cref{fig:framework}b, the NeRF network only contains three SIREN blocks to minimize runtime memory complexity.

Our experiments show that the viewing direction $\vd$ will cause inconsistencies of face identities under multiple views (see~\cref{fig:ablations_xyz_d}a). Thus we do not take $\vd$ as input, which is different from the original NeRF~\cite{mildenhall2020NeRF}. Furthermore, instead of predicting the color $\vc$, we let the NeRF network predict a more general feature $\vv$~\cite{niemeyer2021GIRAFFE}. As a result, the proposed NeRF function is given by
\begin{equation}
  \resizebox{\hsize}{!}{$
      g: \R^{\mathrm{dim}(\vx)} \times \R^{\mathrm{dim}(\vz_s)}  \rightarrow \R^+ \times \R^{\mathrm{dim}(\vv)}, \ (\vx, \vz_s)                                               \mapsto(\sigma, \vv),
    $}
  \label{equ:nerf_feature}
\end{equation}
where $\vv$ is a feature vector corresponding to point $\vx$. $\vz_s$ is a shape code that is shared by all pixels of a generated image.

\begin{figure*}[!t]
  \centering
  \begin{minipage}[t]{0.68\linewidth}
    \begin{center}
      \includegraphics[height=3.4cm]{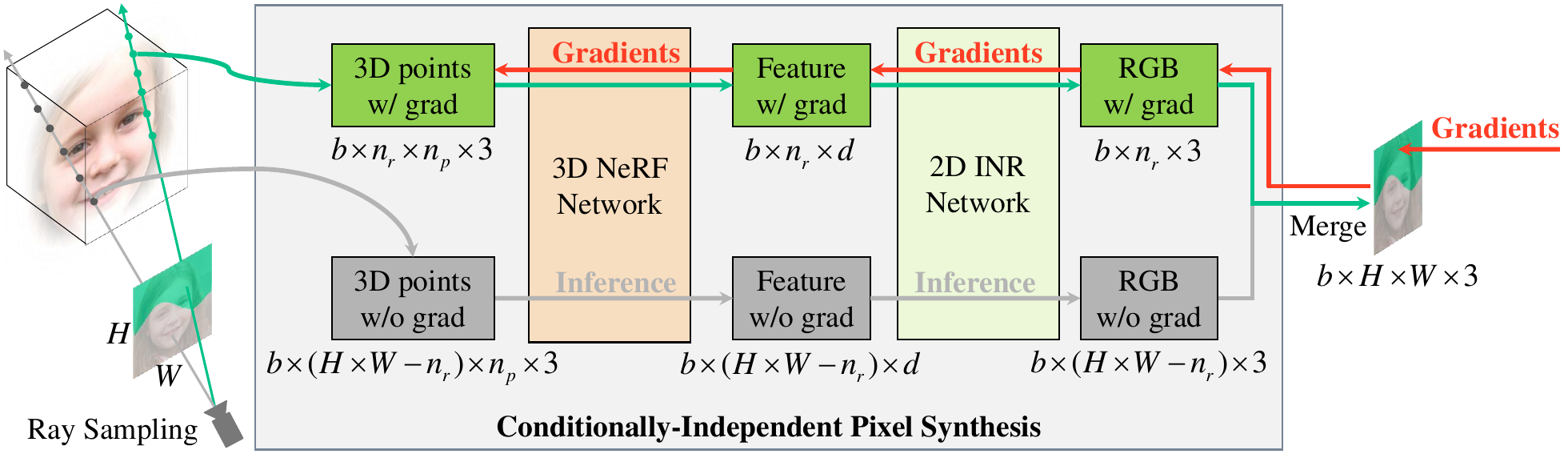}
    \end{center}
    \vspace{-0.6cm}
    \caption{Partial gradient backpropagation. In the training phase, the gradient calculation is turned on during the forward pass only for the green rays sampled randomly. The remaining rays do not participate in the backpropagation (the grey rays).}
    \vspace{-0.4cm}
    \label{fig:partial_grad}
  \end{minipage}%
  \hspace{0.002\linewidth}
  \vline
  \hspace{0.005\linewidth}
  \centering
  \begin{minipage}[t]{0.3\linewidth}
    \centering
    \includegraphics[width=\linewidth]{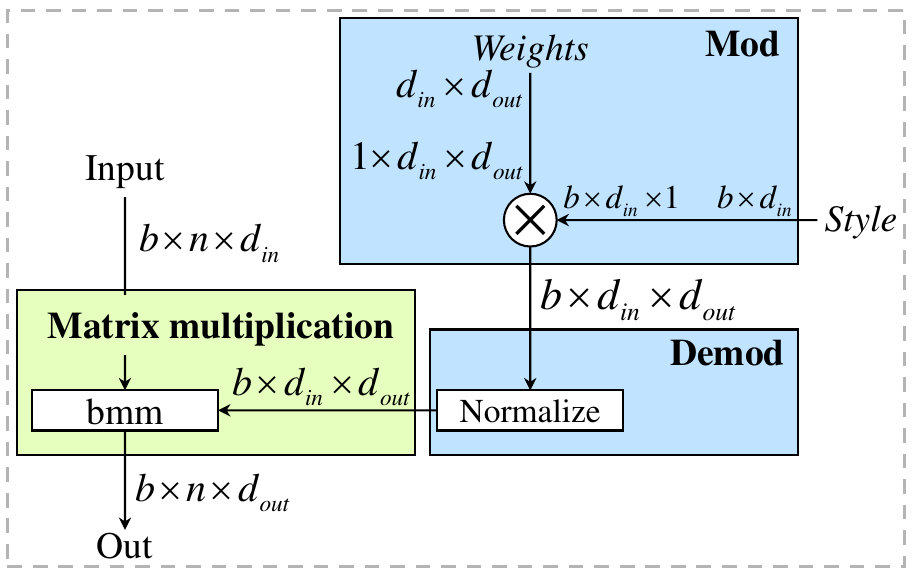}
    \vspace{-0.7cm}
    \caption{An efficient implementation of modulated fully connected layer (ModFC) using $\mathrm{bmm}$.}
    \vspace{-1cm}
    \label{fig:modfc}
  \end{minipage}
\end{figure*}

\paragraph{Volume Rendering}
As described in~\cref{equ:nerf_feature}, the neural radiance field represents a scene as the volume density $\sigma$ and feature vector $\vv$ at any point in space. Let $\vo$ be the camera origin. For each pixel, we cast a ray $\mathbf{r}(t)=\vo+t \vd$ from origin $\vo$ towards the pixel. We sample points along the camera ray $\mathbf{r}(t)$ and transform the 3D coordinates of the points into volume densities and feature vectors using~\cref{equ:nerf_feature}. Using the classical volume rendering~\cite{kajiya1984Ray}, the overall feature vector $V_\mathbf{r}$ corresponding to the ray $\mathbf{r}(t)$ is given by
\begin{equation}
  \resizebox{\hsize}{!}{$
      V_\mathbf{r}  =\int_{t_{n}}^{t_{f}} T(t) \sigma(\mathbf{r}(t)) \vv(\mathbf{r}(t) ) dt, \ T(t)          =\exp \left(-\int_{t_{n}}^{t} \sigma(\mathbf{r}(s)) d s\right),
    $}
  \label{equ:volume_render}
\end{equation}
where $t_n$ and $t_f$ are near and far bounds, respectively.

\subsection{INR Network for Appearance}
\label{sec:deep_inr}

\paragraph{Deep INR Network}
To generate images at $H\times W$ resolution, the NeRF network outputs feature maps of shape $\mathrm{dim}(V_\mathbf{r}) \times H \times W$, where $V_\mathbf{r}$ is the feature vector calculated by~\cref{equ:volume_render}. Next, we need to convert these feature maps into the RGB space. We adopt an implicit neural representation (INR) network, where each pixel value is calculated independently~\cite{anokhin2021Image} given the feature vector $V_\mathbf{r}$. As shown in~\cref{fig:framework}c, the INR network contains nine blocks, and each block contains two modulated fully connected layers (ModFC)~\cite{karras2019Analyzing,anokhin2021Image}. We add a tRGB (\ie, a fully connected layer) layer after each block to convert the intermediate feature maps to RGB values. The final RGB values are the summation of all intermediate RGB values.

\paragraph{Partial Gradient Backpropagation}
\label{sec:partial_gradient}
Our generator network is a columnar structure without any up-sampling or down-sampling operations. Therefore, directly training it on high-resolution images is challenging due to limited GPU memory. Note that the generator synthesizes each pixel value independently. Leveraging this property, we propose a training strategy named \textit{partial gradient backpropagation} for training on high-resolution images.

As shown in Fig.~\ref{fig:partial_grad}, to synthesize an image at $H\times W$ resolution, we randomly sample $n_r$ rays (green rays in Fig.~\ref{fig:partial_grad}), and then convert these rays into $n_r$ RGB values, with the gradient calculation enabled. On the other hand, the remaining $H\times W - n_r$ rays (grey rays in Fig.~\ref{fig:partial_grad}) are converted to RGB values, but the gradient calculation is disabled to save memory. Finally, all the RGB values are combined into a high-resolution image, which will be presented to the discriminator. This strategy allows us to train the generator on high-resolution images. Compared with the patch-based method~\cite{schwarz2020GRAF,anokhin2021Image}, partial gradient backpropagation ensures that the discriminator observes full natural images rather than image patches at low resolution.


\paragraph{Efficient Implementation for ModFC}
Like the NeRF network, the INR network also adopts a style-based architecture. As shown in~\cref{fig:framework}d, a mapping network $m_a: \mathcal{Z}_a \rightarrow \mathcal{W}_a$ turns $\vz_a$ into $\vw_a$, where the stochasticity of appearance comes from the code $\vz_a$. Then, $\vw_a$ is mapped to style vectors using affine layers (\ie, FC layer). The style vectors are injected into the INR network using Modulated Fully Connected (ModFC) layers.

CIPS~\cite{anokhin2021Image} regards ModFC as a special case of $1\times1$ convolutional layer and implements ModFC with the off-the-shelf modulated convolutional layer~\cite{karras2019Analyzing}. The modulated convolution is implemented using grouped convolution, which is not efficient for ModFC. In fact, we can directly utilize the batch matrix multiplication ($\mathrm{bmm}$) to implement ModFC more efficiently. As shown in Fig.~\ref{fig:modfc}, ModFC consists of \textsf{Mod}, \textsf{Demod}, and a batch matrix multiplication operation. Mathematically, let $\mW \in \R^{d_{in} \times d_{out}}$ be the weights of a fully connected layer, $\mS \in \R^{b \times d_{in}}$ be a batch of style vectors, and $\mX \in \R^{b \times n \times d_{in}}$ be the input with $n$ being the length of the sequence. We first resize $\mW$ and $\mS$ to shapes of ${1\times d_{in} \times d_{out}}$ and ${b \times d_{in} \times 1}$, respectively. The \textsf{Mod} operation is given by $\mW^{'}=\mW \otimes \mS$, where $\otimes$ stands for tensor-broadcasting multiplication and $\mW^{'} \in \R^{b \times d_{in} \times d_{out}}$. The \textsf{Demod} operation is given by $\mW^{''}=\mW^{'}\otimes \left( \sum_{d_{in}}(\mW^{'}_{\cdot,d_{in},\cdot})^2 + \epsilon \right)^{-\frac{1}{2}}$, where $\epsilon$ is a small constant and $\mW^{''} \in \R^{b \times d_{in} \times d_{out}}$. Finally, we use $\mW{''}$ to linearly map the input $\mX \in \R^{b \times n \times d_{in}}$ (\ie, $\mY = \mX \times W^{''}$, and $\mY \in \R^{b \times n \times d_{out}}$), which is achieved through the batch matrix multiplication function\footnote{The function is $\mathrm{bmm}(\cdot)$ in PyTorch.}. Experiments substantiate that this implementation is more efficient than the implementation using grouped convolution (see~\cref{fig:params_speed}).

\subsection{Overcoming Mirror Symmetry Issue with Auxiliary Discriminator}
\label{sec:mirror_symmetry}

\begin{figure}[!t]
  \centering
  \begin{minipage}[]{\linewidth}
    \footnotesize
    \centering
    \renewcommand{\tabcolsep}{1pt} \renewcommand{\arraystretch}{1}
    \resizebox{\linewidth}{!}{%
      \begin{tabular}{*{7}{m{0.15\textwidth}}}
        \includegraphics[width=\linewidth]{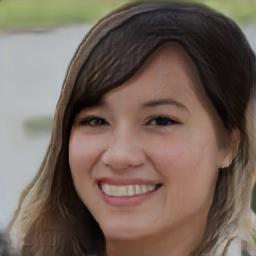} &
        \includegraphics[width=\linewidth]{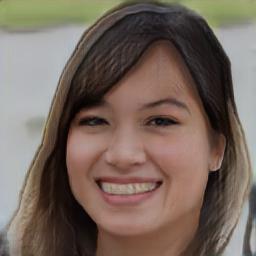} &
        \includegraphics[width=\linewidth]{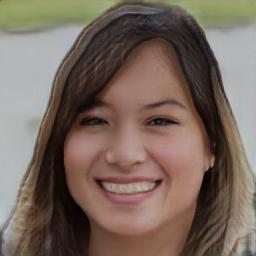} &
        \includegraphics[width=\linewidth]{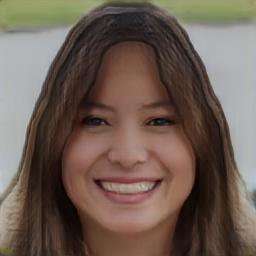} &
        \includegraphics[width=\linewidth]{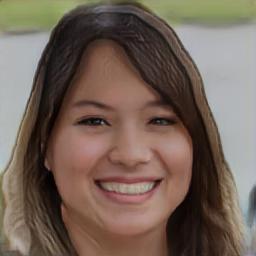} &
        \includegraphics[width=\linewidth]{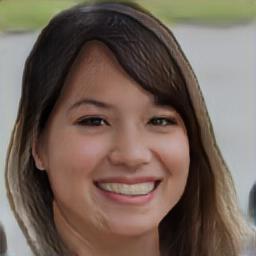} &
        \includegraphics[width=\linewidth]{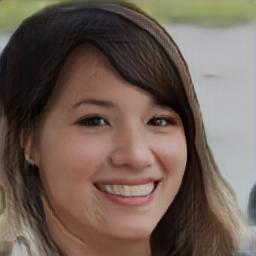}
        \\
        \multicolumn{7}{c}{(a) GIRAFFE~\cite{niemeyer2021GIRAFFE} (mirror symmetry)}
        \\
        \includegraphics[width=\linewidth]{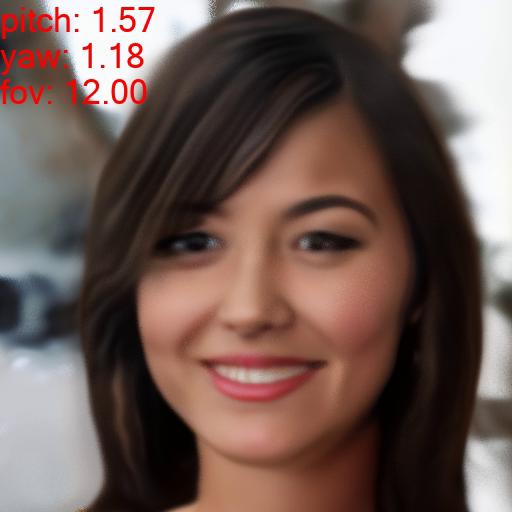}    &
        \includegraphics[width=\linewidth]{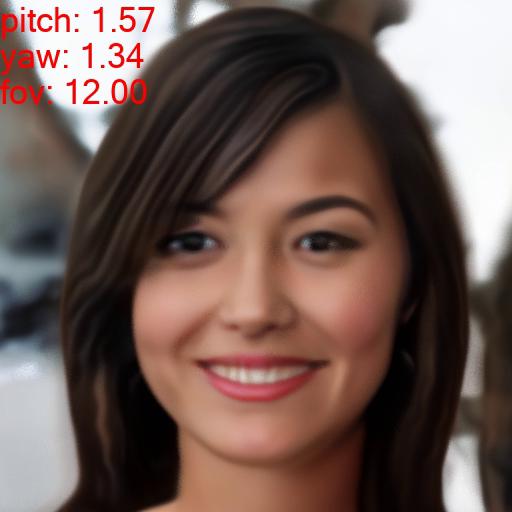}    &
        \includegraphics[width=\linewidth]{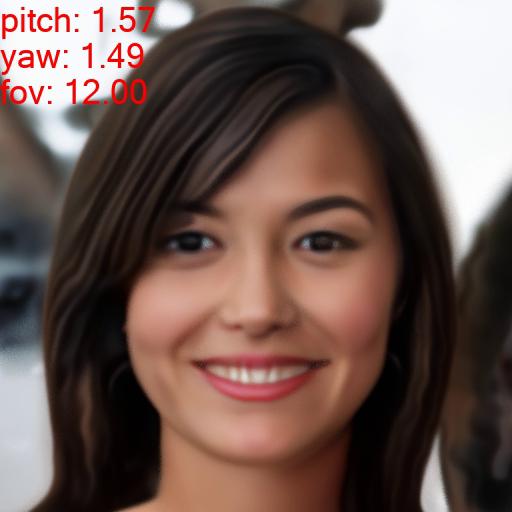}    &
        \includegraphics[width=\linewidth]{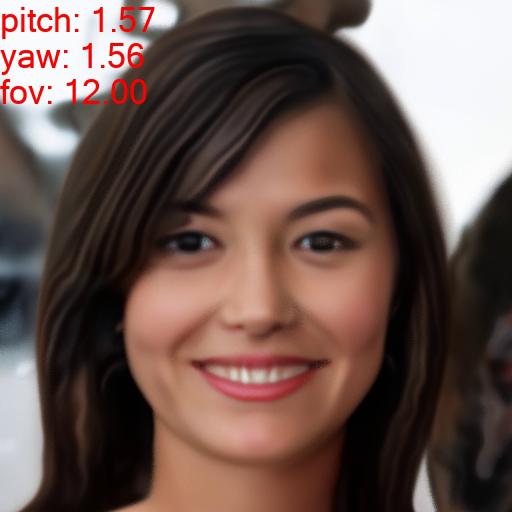}    &
        \includegraphics[width=\linewidth]{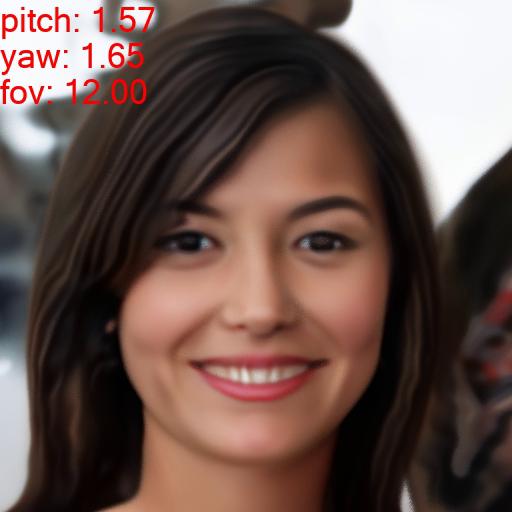}    &
        \includegraphics[width=\linewidth]{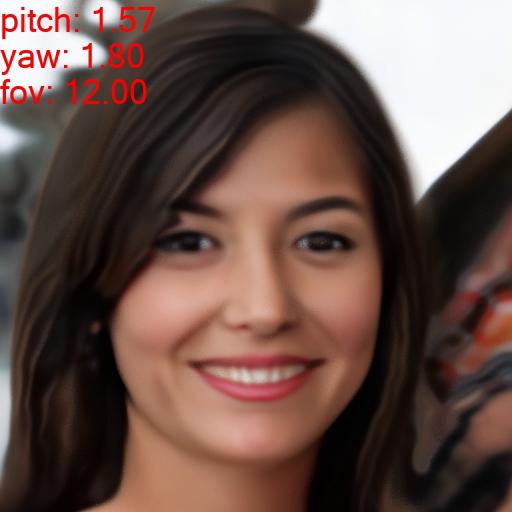}    &
        \includegraphics[width=\linewidth]{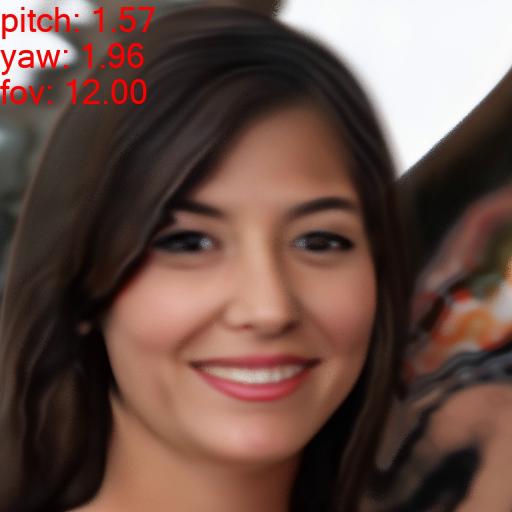}
        \\
        \multicolumn{7}{c}{(b) pi-GAN~\cite{chan2021piGAN} (no mirror symmetry)}
        \\
        \includegraphics[width=\linewidth]{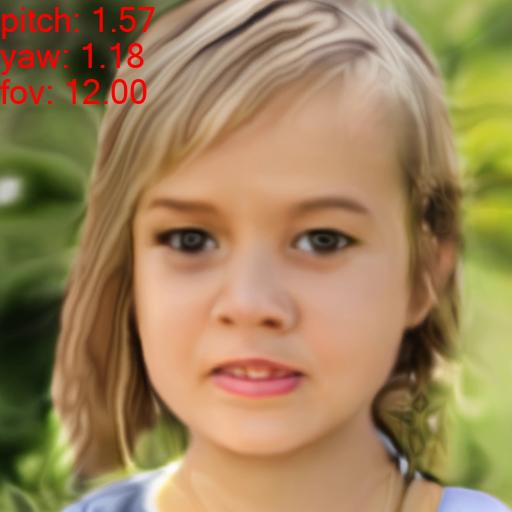}   &
        \includegraphics[width=\linewidth]{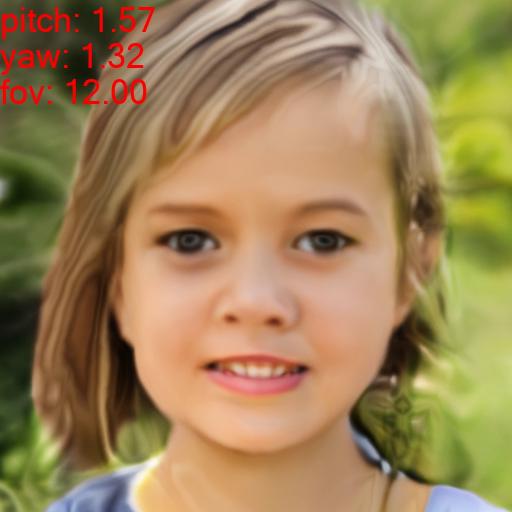}   &
        \includegraphics[width=\linewidth]{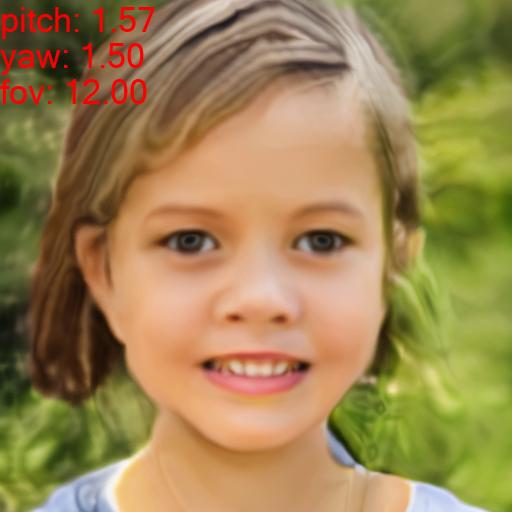}   &
        \includegraphics[width=\linewidth]{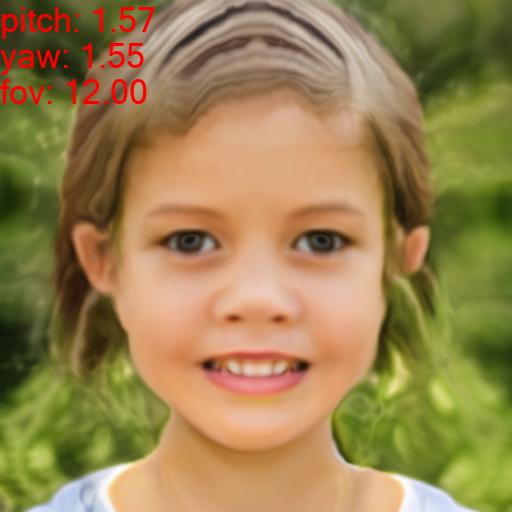}   &
        \includegraphics[width=\linewidth]{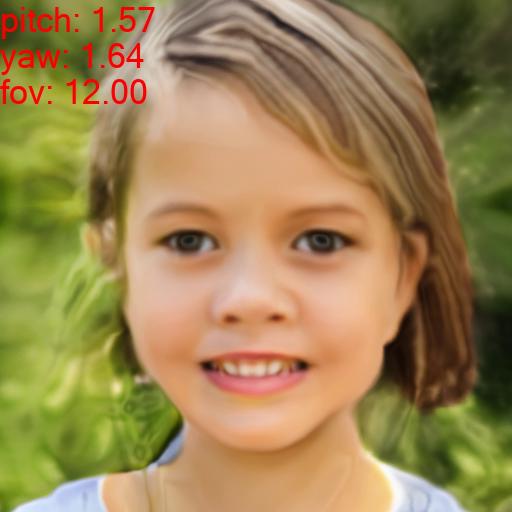}   &
        \includegraphics[width=\linewidth]{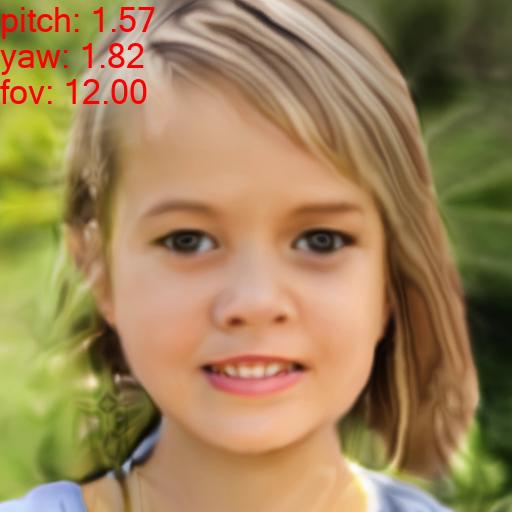}   &
        \includegraphics[width=\linewidth]{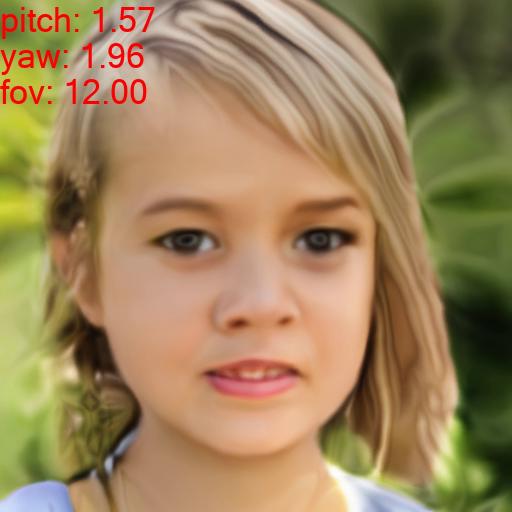}
        \\
        \multicolumn{7}{c}{(c) CIPS-3D w/o auxiliary discriminator (mirror symmetry)}
        \\
        \includegraphics[width=\linewidth]{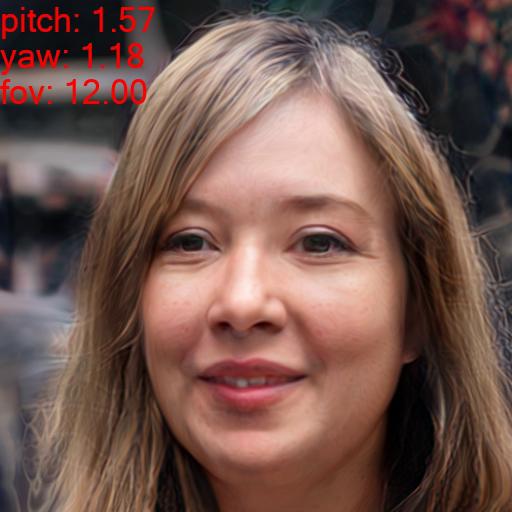}   &
        \includegraphics[width=\linewidth]{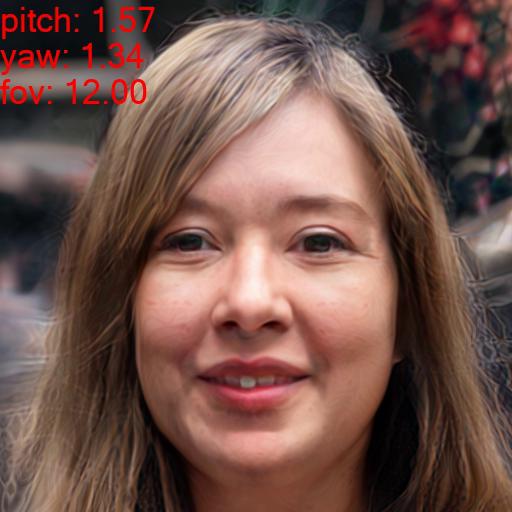}   &
        \includegraphics[width=\linewidth]{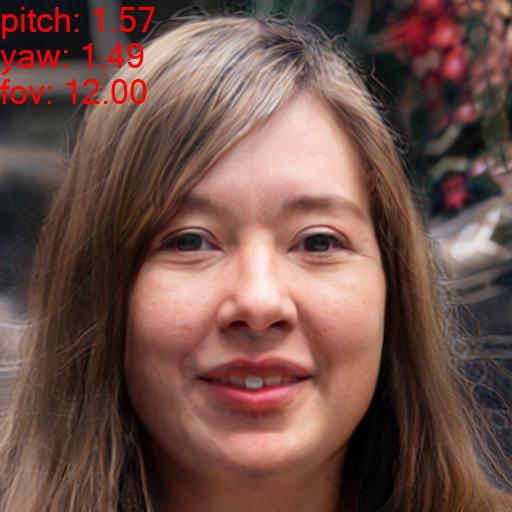}   &
        \includegraphics[width=\linewidth]{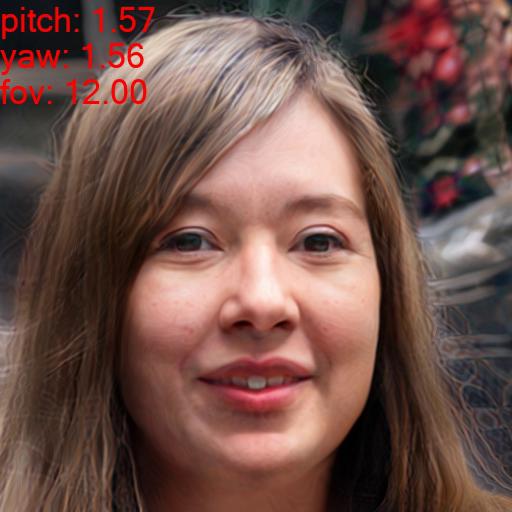}   &
        \includegraphics[width=\linewidth]{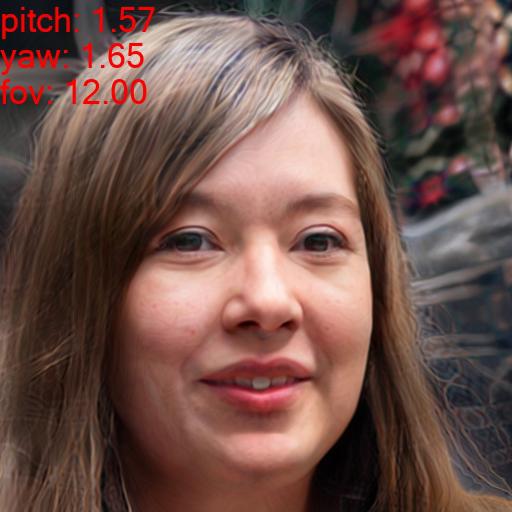}   &
        \includegraphics[width=\linewidth]{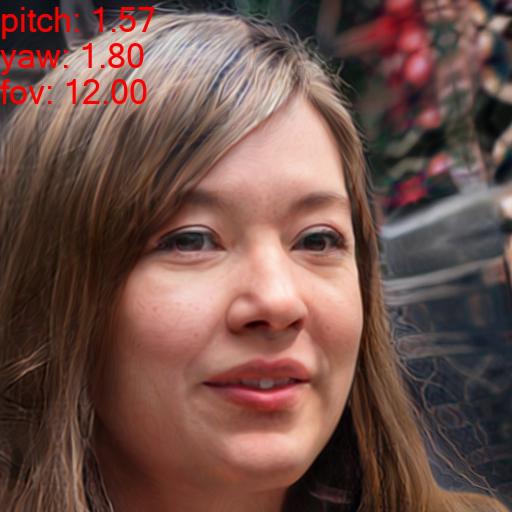}   &
        \includegraphics[width=\linewidth]{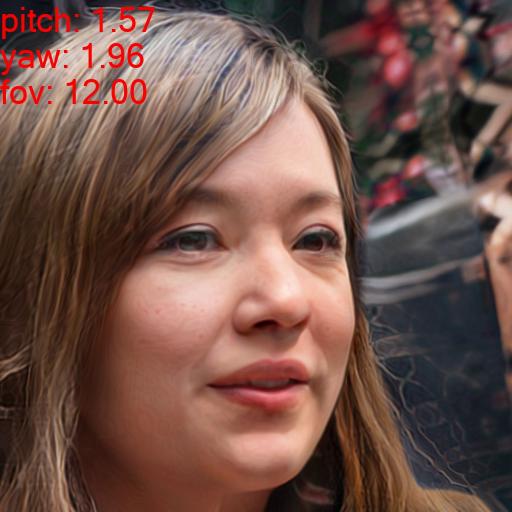}
        \\
        \multicolumn{7}{c}{(d) CIPS-3D w/ auxiliary discriminator (no mirror symmetry)}
      \end{tabular}%
    }
    \vspace{-0.3cm}
    \caption{The images of each row are synthesized with different yaw angles. Mirror symmetry exists in (a) and (c). (d): The auxiliary discriminator helps to overcome the mirror symmetry. }
    \vspace{0.3cm}
    \label{fig:mirror_symm_figs}
  \end{minipage}\\
  \centering
  \begin{minipage}[t]{0.6\linewidth}
    \centering
    \includegraphics[height=2.6cm]{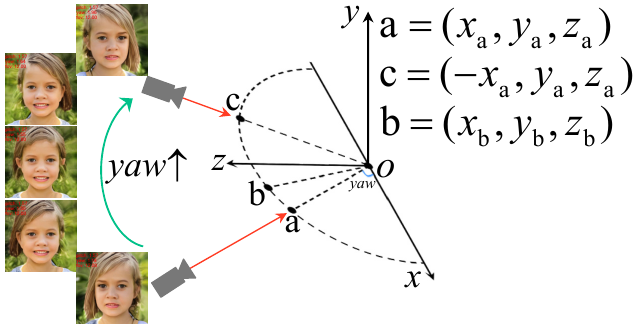}
    \vspace{-0.3cm}
    \caption{The direction of the bangs changes suddenly near the yaw angle of $\frac{\pi}{2}$. Please zoom in to see the yaw angles (at the upper left corner of the images).}
    \label{fig:mirror_symm_coord}
  \end{minipage}
  \vline
  \vspace{0.01\linewidth}
  \centering
  \begin{minipage}[t]{0.37\linewidth}
    \centering
    \includegraphics[width=\linewidth]{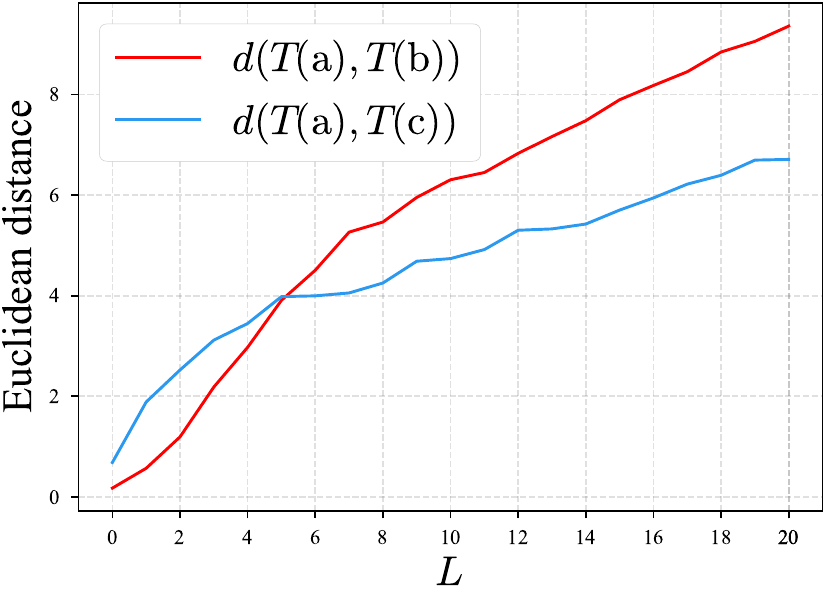}
    \vspace{-0.7cm}
    \caption{As $L$ increases, the distance between $\mathrm{a}$ and its symmetry point $\mathrm{c}$ is less than the distance between $\mathrm{a}$ and its neighbor $\mathrm{b}$. }
    \label{fig:pos_encoding}
  \end{minipage}
\end{figure}


In practice, we found a mirror symmetry problem for the generator composed of the NeRF network and the INR network. As shown in Fig.~\ref{fig:mirror_symm_figs}c and Fig.~\ref{fig:mirror_symm_coord}, the direction of the bangs changes suddenly near the yaw angle of $\frac{\pi}{2}$. Interestingly, GIRAFFE~\cite{niemeyer2021GIRAFFE}, composed of a deeper NeRF network and a CNN decoder, also has this disturbing problem (see Fig.~\ref{fig:mirror_symm_figs}a). We identify two sources for the mirror symmetry: (i) the positional encoding function~\cite{mildenhall2020NeRF}, and (ii) the mirror symmetry of the input coordinates of the NeRF network.

The positional encoding function $\gamma: \R \rightarrow \R^{2L}$, mapping a scalar to a high-dimensional vector, is given by
\begin{equation}
  \resizebox{\hsize}{!}{$
      \gamma(t; L)=
      \left(\sin \left(2^{0} t \pi\right), \cos \left(2^{0} t \pi\right), \ldots, \sin \left(2^{L-1} t \pi\right), \cos \left(2^{L-1} t \pi\right)\right),
    $}
  \label{equ:position_encoding}
\end{equation}
where $t$ is a scalar, and $L$ is a hyperparameter determining the dimension of the mapping space.

\textbf{Definition 1.}
Let $(X, d_x)$ and $(Y, d_y)$ be two metric spaces. A mapping function $f:X \rightarrow Y$ is called distance preserving if for any $a, b \in X$, one has
\begin{equation}
  d_{y}(f(a), f(b))=d_{x}(a, b).
  \label{equ:distance_preserving}
\end{equation}

\textbf{Proposition 1.} A positional encoding function $T: \R^3 \rightarrow \R^{3+6L}$ is given by
\begin{equation}
  T(x, y, z; L)=(x, y, z, \gamma(x;L), \gamma(y;L), \gamma(z;L)),
  \label{equ:proposition}
\end{equation}
where $\gamma(\cdot; L)$ is defined by~\cref{equ:position_encoding}. Then $T(\cdot; L=10)$ \footnote{$T(\cdot; L=10)$ is often adopted by NeRF-based networks.} is not distance preserving.

\textit{\textbf{Proof.}}\quad
As shown in Fig.~\ref{fig:mirror_symm_coord}, let $\mathrm{a}=(\cos\frac{70\pi}{180}, 0, \sin\frac{70\pi}{180})$, $\mathrm{c}=(-\cos\frac{70\pi}{180}, 0, \sin\frac{70\pi}{180})$, and $\mathrm{b}=(\cos\frac{80\pi}{180}, 0, \sin\frac{80\pi}{180})$, then we have
\begin{equation}
  d(\mathrm{a}, \mathrm{b}) < d(\mathrm{a}, \mathrm{c}),
  \label{equ:d_ab_lt_d_ac}
\end{equation}
where $d$ represents Euclidean distance, and $\mathrm{a}$ and $\mathrm{c}$ are symmetric with respect to the $yOz$ plane. We apply $T(\cdot; L)$ to $\mathrm{a}$, $\mathrm{b}$, and $\mathrm{c}$ respectively, and draw the distance between them in Fig.~\ref{fig:pos_encoding}. Then we know that
\begin{equation}
  \resizebox{\hsize}{!}{$
      d\big( T\left(\mathrm{a}; L=10\right), T\left(\mathrm{b}; L=10\right) \big) > d\big( T\left(\mathrm{a}; L=10\right), T\left(\mathrm{c}; L=10\right) \big).
    $}
  \label{equ:dT_ab_gt_dT_ac}
\end{equation}
Supposing that $T(\cdot; L=10)$ is distance preserving, according to~\cref{equ:distance_preserving} and~\cref{equ:d_ab_lt_d_ac}, we get
\begin{equation}
  \resizebox{\hsize}{!}{$
      d\big( T\left(\mathrm{a}; L=10\right), T\left(\mathrm{b}; L=10\right) \big) < d\big( T\left(\mathrm{a}; L=10\right), T\left(\mathrm{c}; L=10\right) \big),
    $}
\end{equation}
which contradicts the fact of ~\cref{equ:dT_ab_gt_dT_ac}. Thus $T(\cdot; L=10)$ is not distance preserving and we finish the proof.
\hfill$\square$

\cref{equ:dT_ab_gt_dT_ac} shows that after positional encoding, the distance between $\mathrm{a}$ and its symmetry point $\mathrm{c}$ is less than the distance between $\mathrm{a}$ and its neighbor $\mathrm{b}$. This may cause the network to predict similar appearance for $\mathrm{a}$ and $\mathrm{c}$, causing mirror symmetry. Therefore, we discard the fixed positional encoding function and adopt a learnable strategy, \ie, the coordinates $(x, y, z)$ are mapped to a high-dimensional space by a fully connected layer followed by a $\mathrm{sine}$ activation~\cite{sitzmann2020Implicit}. However, mirror symmetry remains.

After closer inspection, we found that the essence of mirror symmetry lies in the symmetry of the coordinate system. As shown in Fig.~\ref{fig:mirror_symm_coord}, the coordinates of $\mathrm{a}$ and $\mathrm{c}$ are almost the same except that the $x$ coordinate differs by a minus sign. The network tends to leverage the symmetry of coordinates to learn a symmetrical appearance. Note that the symmetry is a double-edged sword, and it facilitates the fitting of symmetrical objects~\cite{wu2020Unsupervised,pan20212D} such as human faces, cat faces, cars, \etc.

We notice that pi-GAN~\cite{chan2021piGAN}, whose generator is a pure NeRF network, does not suffer from mirror symmetry (see Fig.~\ref{fig:mirror_symm_figs}b). This substantiates that the discriminator can prevent the NeRF network from falling into the pitfall of mirror symmetry. However, both GIRAFFE's generator (NeRF + CNN decoder) and our generator (NeRF + INR network) suffer from mirror symmetry. It turns out that the discriminator cannot effectively regularize the NeRF network when there is a 2D network between the 3D NeRF network and the discriminator. Therefore, we propose to utilize an auxiliary discriminator to supervise the output of the NeRF network directly. In particular, the output of the NeRF network is mapped to the RGB space by a fully connected layer (see Fig.~\ref{fig:framework}b), and the RGB is presented to the auxiliary discriminator. As shown in Fig.~\ref{fig:mirror_symm_figs}d, the mirror symmetry disappeared after applying the auxiliary discriminator.



\section{Experiments}

\subsection{Implementation details}

During training, we let the virtual camera be located on the surface of a unit sphere and look at the origin. The pitch and yaw are randomly sampled from predefined distributions, respectively. Both the main discriminator and the auxiliary discriminator adopt the StyleGAN2~\cite{karras2019Analyzing} discriminator architecture, but the auxiliary discriminator has fewer channels. The model is trained with a standard non-saturating logistic GAN loss with R1 penalty~\cite{mescheder2018Which}. We adopt Adam~\cite{kingma2014Adam} to train the networks with $\beta_0=0$ and $\beta_1=0.999$. Following pi-GAN~\cite{chan2021piGAN}, we adopt a progressive training strategy where we start training at $64^2$ resolution and progressively increase the resolution up to $512^2$. Note that the generator architecture remains fixed, but the resolution of the generator is increased by sampling more rays. We train the model with eight Tesla V100 GPUs and batch size of $32$. When the resolution of the generator reaches $512\times512$, the number of rays participating in the gradient calculation is set to $400^2$ to save GPU VRAM.

\begin{table}[t]
  \caption{Comparison with SOTA on FFHQ. We computed FID and KID ($\times 10^3$) between 50k generated images and all training images using the torch-fidelity library~\cite{obukhov2020torchfidelity}. $\dagger$ stands for quoting from the paper. CIPS-3D achieves the state-of-the-art for 3D-aware GANs, surpassing the very recent method StyleNeRF by clear margins. }
  \vspace{-0.3cm}
  \label{tab:fid_results}
  \centering
  \resizebox{0.9\linewidth}{!}{%
    \centering
    \begin{tabular}{llcccc}
      \toprule
                           & \multicolumn{1}{l}{\multirow{2}{*}{Methods}} & \multicolumn{2}{c|}{$256\times256$}           & \multicolumn{2}{c}{$1024\times1024$}
      \\ \cmidrule{3-6}
      \multicolumn{2}{c}{} & $\mathrm{FID}\downarrow$                     & \multicolumn{1}{c|}{$\mathrm{KID}\downarrow$} & $\mathrm{FID}\downarrow$             & $\mathrm{KID}\downarrow$                   \\
      \midrule
      \multirow{2}{*}{2D}  & StyleGAN2~\cite{karras2019Analyzing}         & $\mathbf{4.30}$                               & $\mathbf{1.07}$                      & $\mathbf{2.86}$          & $\mathbf{0.53}$ \\
                           & CIPS~\cite{anokhin2021Image}                 & $23.06$                                       & $23.04$                              & $10.03$                  & $4.79$          \\
      \midrule
      \multirow{4}{*}{3D}  & GIRAFFE~\cite{niemeyer2021GIRAFFE}           & $63.33$                                       & $50.94$                              & -                        & -               \\
                           & pi-GAN~\cite{chan2021piGAN}                  & $34.56$                                       & $26.58$                              & $35.97$                  & $28.09$         \\
                           & StyleNeRF~\cite{anonymous2021StyleNeRF}      & $8.00$$^\dagger$                              & $3.70$$^\dagger$                     & -                        & -               \\
                           & CIPS-3D (ours)                               & $\mathbf{6.97}$                               & $\mathbf{2.87}$                      & $\mathbf{12.26}$         & $\mathbf{7.74}$ \\
      \bottomrule
    \end{tabular}
  }
  \vspace{-0.3cm}
\end{table}

\begin{figure}[!t]
  \begin{subfigure}[]{0.49\linewidth}
    \centering
    \captionsetup{justification=centering,margin=0cm}
    \includegraphics[width=\linewidth]{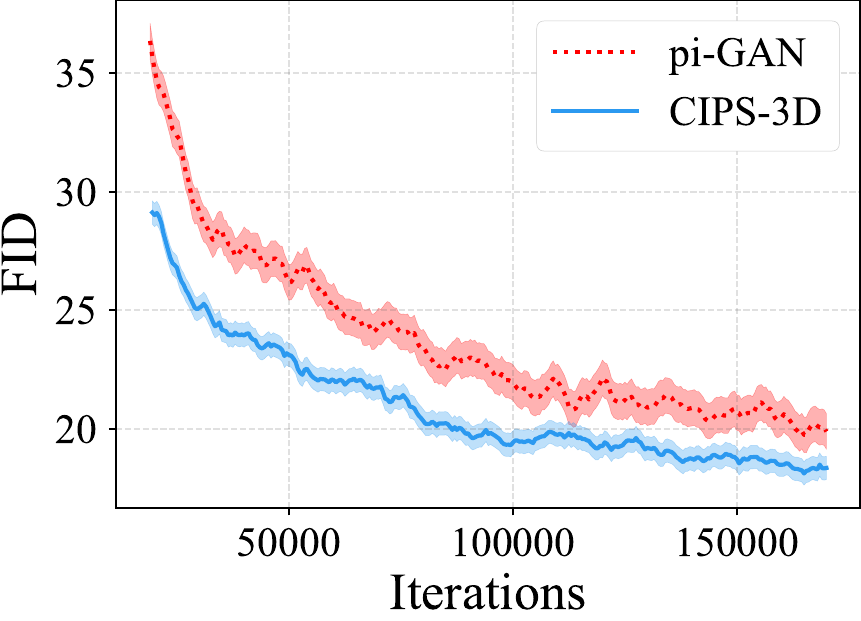}
    \caption{Training at $64^2$ resolution.}
    \label{fig:sub:progressive_64}
  \end{subfigure}%
  \hspace{0.01\linewidth}
  \begin{subfigure}[]{0.49\linewidth}
    \centering
    \captionsetup{justification=centering,margin=0cm}
    \includegraphics[width=\linewidth]{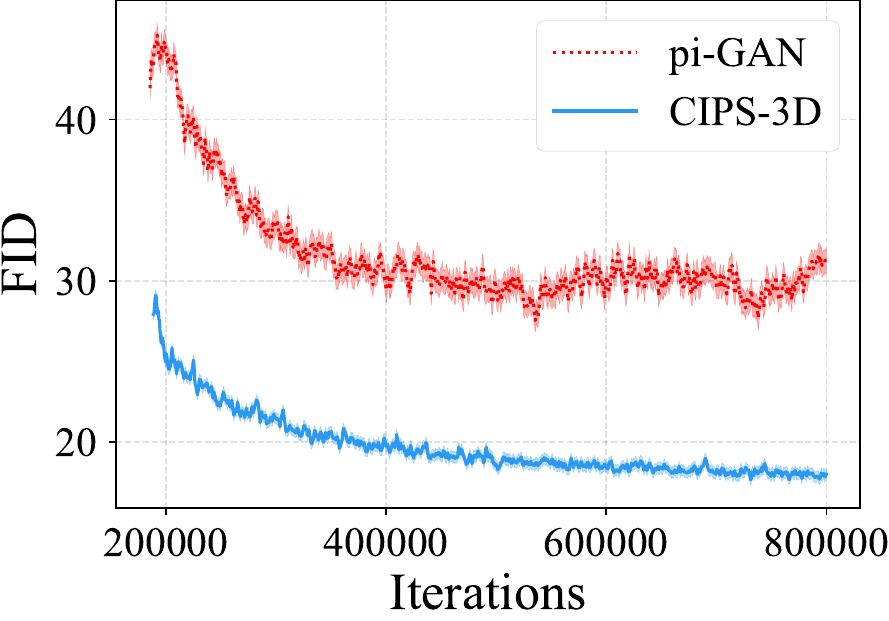}
    \caption{Training at $128^2$ resolution.}
    \label{fig:sub:progressive_128}
  \end{subfigure}
  \vspace{-0.3cm}
  \caption{Progressive training on FFHQ. Generators are initially trained at $64^2$ resolution (a), then at $128^2$ resolution (b). pi-GAN converges steadily at low resolution, but deteriorates at higher resolution. For efficiency of training, FID is calculated between $2048$ generated images and $2048$ real images.}
  \label{fig:progressive_training}
  \vspace{-0.3cm}
\end{figure}

\begin{figure}[!t]
  \centering
  \includegraphics[width=0.65\linewidth]{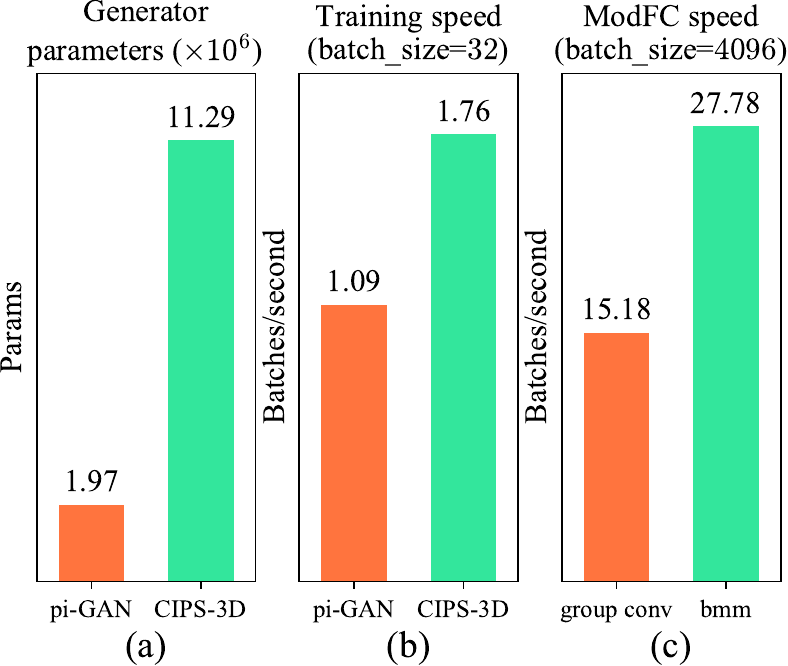}
  \vspace{-0.3cm}
  \caption{(a) and (b): CIPS-3D has more generator parameters, but its training speed is higher than that of pi-GAN which adopts a pure NeRF generator. (c): Compared with group conv, $\mathrm{bmm}$ improves the speed of ModFC considerably. Please refer to \cref{sec:exp:evaluation} for details.}
  \label{fig:params_speed}
  \vspace{-0.5cm}
\end{figure}

\subsection{Evaluation}
\label{sec:exp:evaluation}

\paragraph{Comparison with SOTA}
We evaluate image quality using Fréchet Inception Distance (FID)~\cite{heusel2017GANs} and Kernel Inception Distance (KID)~\cite{binkowski2021Demystifying}. The baseline models contain the current start-of-the-art 3D-aware GANs: GIRAFFE~\cite{niemeyer2021GIRAFFE}, pi-GAN~\cite{chan2021piGAN}, and StyleNeRF~\cite{anonymous2021StyleNeRF}. We also present the results of 2D GANs, such as StyleGAN2~\cite{karras2019Analyzing} and CIPS~\cite{anokhin2021Image}, for your reference. As shown in \cref{tab:fid_results}, our method sets new records for 3D-aware GANs on FFHQ~\cite{karras2019StyleBased} with impressive FID scores of $6.97$ and $12.26$ for images at $256^2$ and $1024^2$ resolution, respectively. Note that our method even outperforms the StyleNeRF~\cite{anonymous2021StyleNeRF} proposed very recently, in terms of FID ($6.97$ \textit{vs.} $8.00$) and KID ($2.87$ \textit{vs.} $3.70$) at $256^2$ resolution. However, our method is still inferior to the 2D state of the art, StyleGAN2, leaving room for improvement in future work.
We present some generated images in~\cref{fig:first_page_fig}. Compared to other 3D-aware GANs, CIPS-3D generates images with sharper details.

\paragraph{Progressively Growing Training}
\cref{fig:progressive_training} shows the FID curves over the course of progressive training. CIPS-3D is clearly better than pi-GAN, especially at higher resolution. Note that it is not easy to train pi-GAN at high resolution. First of all, the generator of pi-GAN is a pure NeRF architecture being memory-intensive. Secondly, as shown in~\cref{fig:sub:progressive_128}, the convergence of pi-GAN has already deteriorated in the middle resolution (\ie, $128^2$). Additionally, we found that pi-GAN is sensitive to hyperparameters, and increasing the depth and capacity of the generator may cause pi-GAN to fail to converge. CIPS-3D circumvents these difficulties by designing a new generator architecture consisting of a shallow 3D NeRF network and a deep 2D INR network.


\paragraph{Efficiency Comparison}
As shown in Fig.~\ref{fig:params_speed}a and \ref{fig:params_speed}b, CIPS-3D has more parameters but is more efficient than pi-GAN that adopts a pure NeRF generator. To increase the number of parameters, we increase the depth of the INR network ($18$ layers) and the dimension of the fully connected layers ($512$ dimensions). Although there are more layers, the training speed of CIPS-3D is still higher than that of pi-GAN because the 2D INR network is inherently higher than the 3D NeRF network in terms of training efficiency. \cref{fig:params_speed}c compares the speed of ModFC using different implementation methods. Compared with group conv, $\mathrm{bmm}$ improves the speed of ModFC considerably (\ie, from $15.18$ batches/second to $27.78$ batches/second, with batch size of $4096$, input/output dimension of $512$). We evaluate the speed $1000$ times and report the average runtime.

\subsection{Ablation Studies}
\label{sec:exp:ablation}

\begin{figure}[!t]
  \centering
  \begin{minipage}[]{\linewidth}
    \captionof{table}{Ablation studies on FFHQ at $64^2$ resolution. We also present the images generated by each compared method in \cref{fig:ablations_xyz_d}a, \cref{fig:ablations_xyz_d}b, \cref{fig:mirror_symm_figs}c, and \cref{fig:mirror_symm_figs}d, respectively. The proposed auxiliary discriminator eliminates the problem of mirror symmetry. FID is calculated between $2048$ generated images and $2048$ real images. Please refer to \cref{sec:exp:ablation} for details.}
    \vspace{-0.3cm}
    \label{tab:ablations}
    \centering
    \resizebox{\linewidth}{!}{%
      \begin{tabular}{ccccc|cc}
        \toprule
        \multicolumn{2}{c|}{Input} & \multicolumn{2}{c}{Positional Encoding} & \multirowcell{2}{Aux                                                                             \\ Disc} & \multirowcell{2}{Mirror                             \\ Symmetry} & \multirow{2}{*}{FID}      \\ \cmidrule{1-4}
        $(x, y, z)$                & \multicolumn{1}{c|}{$\vd$}              & Fixed Function       & $\mathrm{FC} + \mathrm{sine}$ &                &                &         \\ 
        \midrule
        \CheckmarkBold             & \CheckmarkBold                          & \CheckmarkBold       &                               &                & \CheckmarkBold & $17.46$
        \\ 
        \CheckmarkBold             &                                         & \CheckmarkBold       &                               &                & \CheckmarkBold & $16.25$
        \\
        \CheckmarkBold             &                                         &                      & \CheckmarkBold                &                & \CheckmarkBold & $18.77$
        \\
        \CheckmarkBold             &                                         &                      & \CheckmarkBold                & \CheckmarkBold &                & $16.46$
        \\
        \bottomrule
      \end{tabular}
    }
  \end{minipage}\\
  \centering
  \begin{minipage}[]{\linewidth}
    \footnotesize
    \centering
    \renewcommand{\tabcolsep}{0.5pt} \renewcommand{\arraystretch}{1}
    \resizebox{\linewidth}{!}{%
      \begin{tabular}{*{7}{m{0.15\textwidth}}}
        \includegraphics[width=\linewidth]{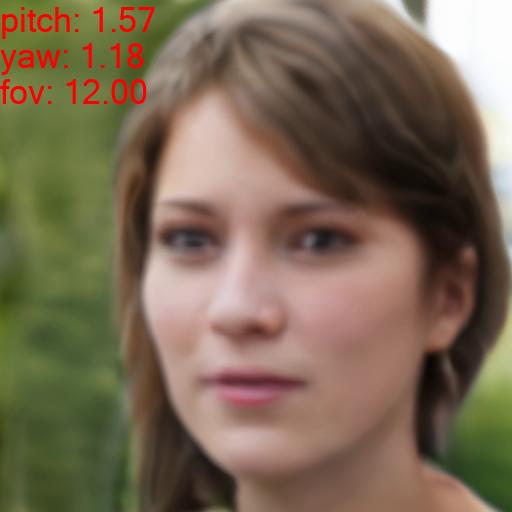} &
        \includegraphics[width=\linewidth]{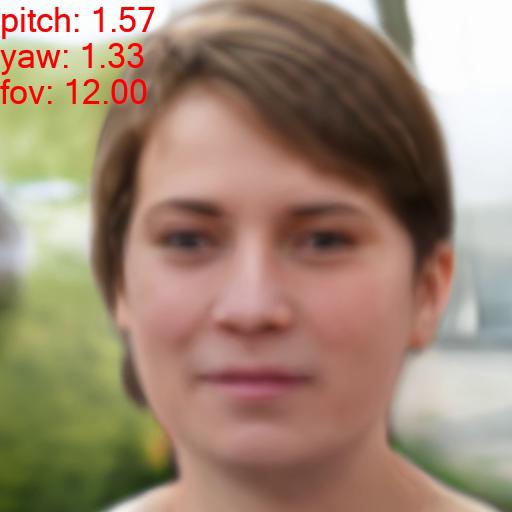} &
        \includegraphics[width=\linewidth]{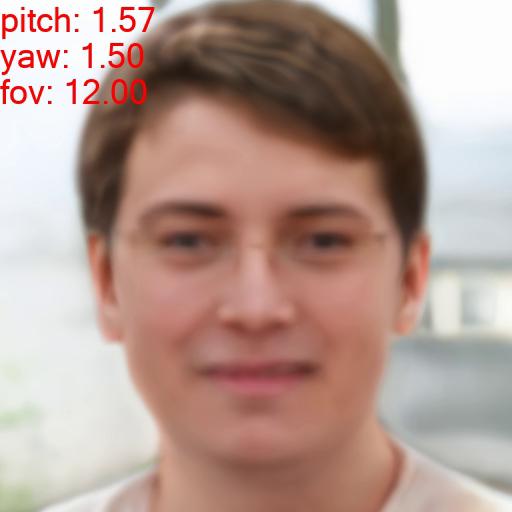} &
        \includegraphics[width=\linewidth]{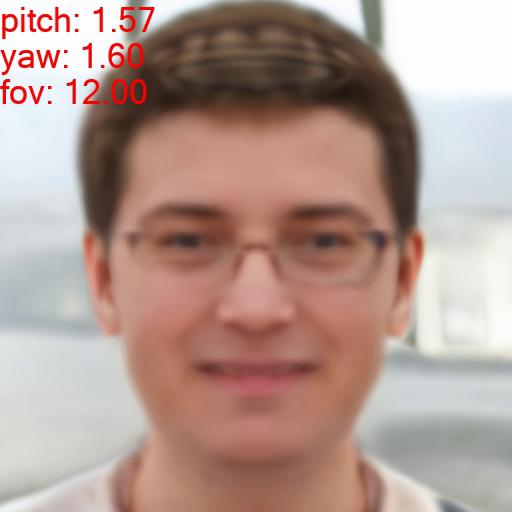} &
        \includegraphics[width=\linewidth]{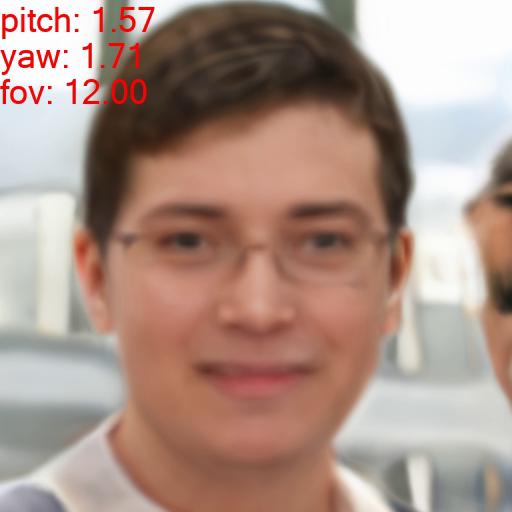} &
        \includegraphics[width=\linewidth]{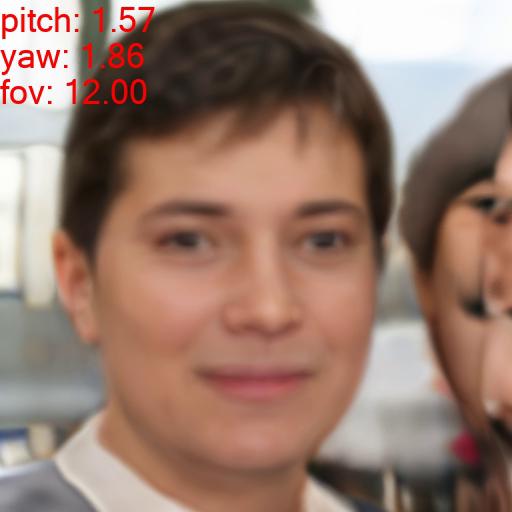} &
        \includegraphics[width=\linewidth]{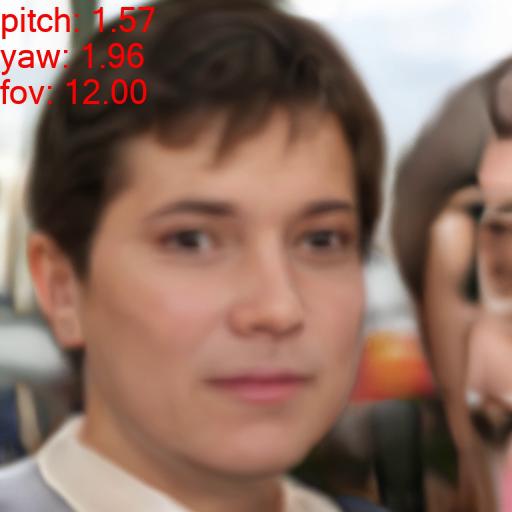}
        \\
        \multicolumn{7}{c}{(a) Ablation study: $(x, y, z) + \vd + \text{Fixed PE function}$}
        \\
        \includegraphics[width=\linewidth]{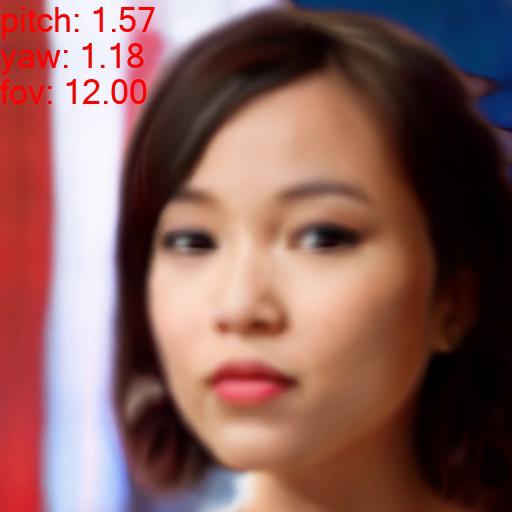}   &
        \includegraphics[width=\linewidth]{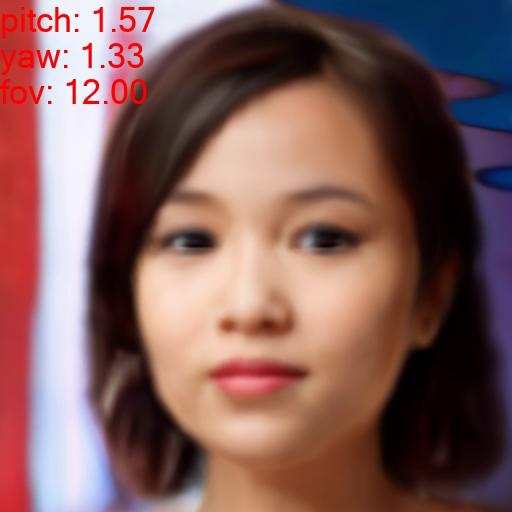}   &
        \includegraphics[width=\linewidth]{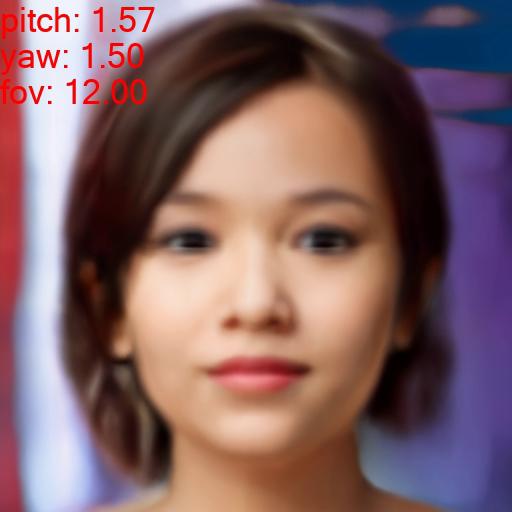}   &
        \includegraphics[width=\linewidth]{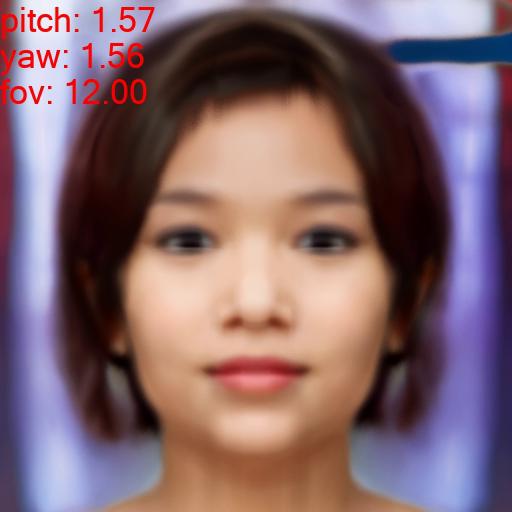}   &
        \includegraphics[width=\linewidth]{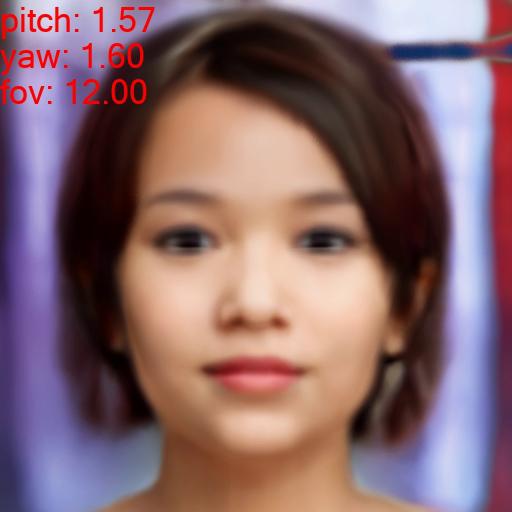}   &
        \includegraphics[width=\linewidth]{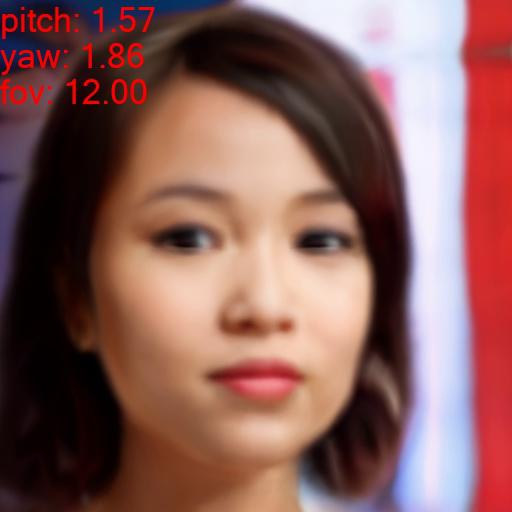}   &
        \includegraphics[width=\linewidth]{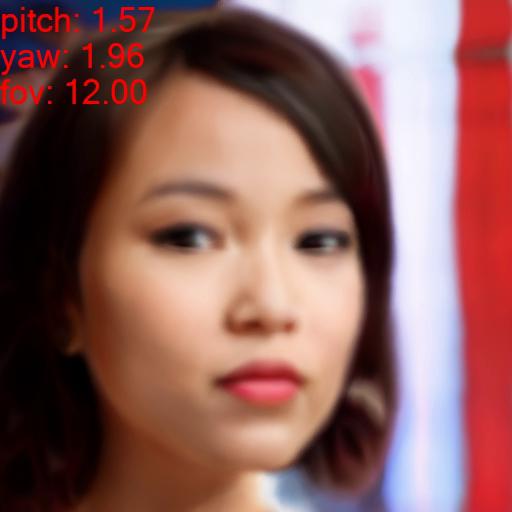}
        \\
        \multicolumn{7}{c}{(b) Ablation study: $(x, y, z) + \text{Fixed PE function}$}
      \end{tabular}
    }
    \vspace{-0.3cm}
    \caption{These images are generated by the ablation methods of the first two rows of \cref{tab:ablations}. The images of each row are synthesized with different yaw angles. Please zoom in to see the pitch and yaw angles (at the upper left corner of the images). (a): Incorporating the viewing direction $\vd$ as input leads to inconsistencies in face identity (please compare the leftmost and rightmost images). Interestingly, StyleNeRF~\cite{anonymous2021StyleNeRF} claimed that the mirror symmetry is caused by the viewing direction $\vd$. However, in our experiments, both (a) and (b) suffer from the mirror symmetry issue, indicating that the viewing direction $\vd$ is not the root cause of the mirror symmetry. PE: positional encoding.}
    \vspace{-0.3cm}
    \label{fig:ablations_xyz_d}
  \end{minipage}
\end{figure}

We conduct ablation studies to help understand the individual components. As shown in the first two rows of \cref{tab:ablations}, discarding the viewing direction $\vd$ for the NeRF network improves FID moderately. \cref{fig:ablations_xyz_d} shows the images generated by these two comparison methods. We deliver two messages. First, incorporating the viewing direction $\vd$ as input leads to inconsistencies in face identity (Fig.~\ref{fig:ablations_xyz_d}a). Second, the generator will suffer from the mirror symmetry issue regardless of whether the viewing direction $\vd$ is used as input or not. We have explained the reason for the mirror symmetry in \cref{sec:mirror_symmetry} and think it is due to the inherent symmetry of the coordinate system.

The second and third rows of \cref{tab:ablations} indicate that the learnable positional encoding function (FC + $\mathrm{sine}$) deteriorates FID compared to the commonly used fixed positional encoding function. After adopting the auxiliary discriminator, the problem of mirror symmetry is solved, and the FID is improved as well (the third row \textit{vs.} the fourth row in \cref{tab:ablations}). Since the fixed positional encoding function (\ie, \cref{equ:proposition}) is not distance preserving, it theoretically may strengthen the problem of mirror symmetry, as analyzed in~\cref{sec:mirror_symmetry}. Thus we decided to adopt the learnable positional encoding function. Our final method includes using only coordinates as input, a learnable positional encoding function, and an auxiliary discriminator.

\begin{figure}[!t]
  \centering
  \includegraphics[width=0.65\linewidth]{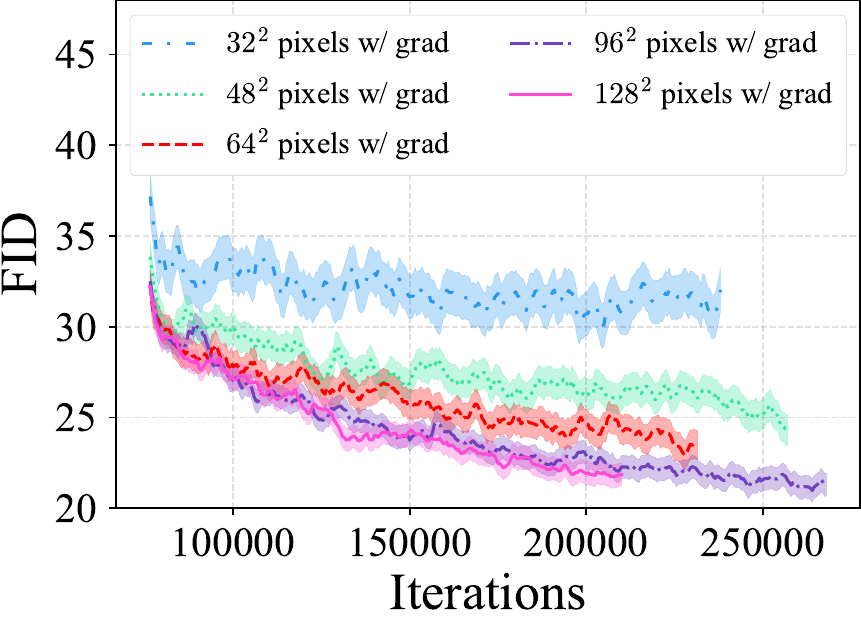}
  \vspace{-0.3cm}
  \caption{Ablations for partial gradient computation on FFHQ at $128^2$ resolution. Too few pixels for the gradient backpropagation during training harms the performance. Using more pixels will narrow the performance gap with using all pixels.}
  \vspace{-0.4cm}
  \label{fig:grad_points}
\end{figure}

Next, we study the effect of the number of pixels participating in calculating the gradient on performance. \cref{fig:grad_points} plots the FID curves over the course of training. Ablations were conducted at $128^2$ resolution on FFHQ. It is evident that too few pixels participating in gradient backpropagation during training will harm the performance. When the number of pixels for the gradient backpropagation reaches $96^2$, the performance is comparable to that of using all pixels (\ie, $128^2$). In addition, we experimentally found that using a small number of pixels (\eg, $48^2$) can still achieve the performance of using all pixels, but at the cost of more training iterations.

\begin{figure}[!t]
  \footnotesize
  \centering
  \renewcommand{\tabcolsep}{5pt} \renewcommand{\arraystretch}{0}
  \resizebox{\linewidth}{!}{%
    \begin{tabular}{ccc|ccc}
      \includegraphics[width=\linewidth]{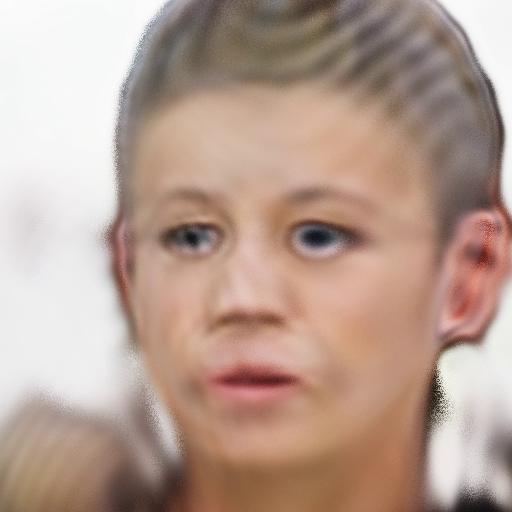} &
      \includegraphics[width=\linewidth]{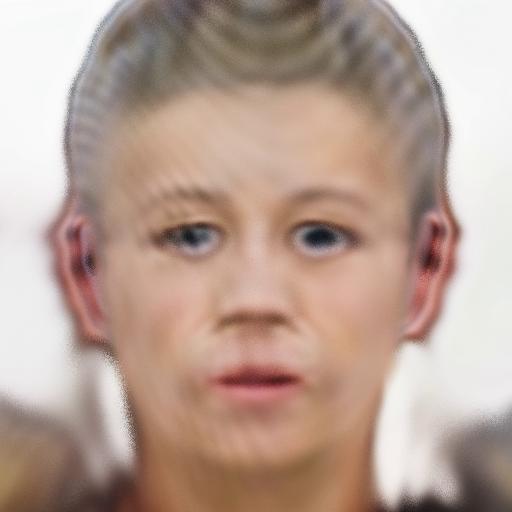} &
      \includegraphics[width=\linewidth]{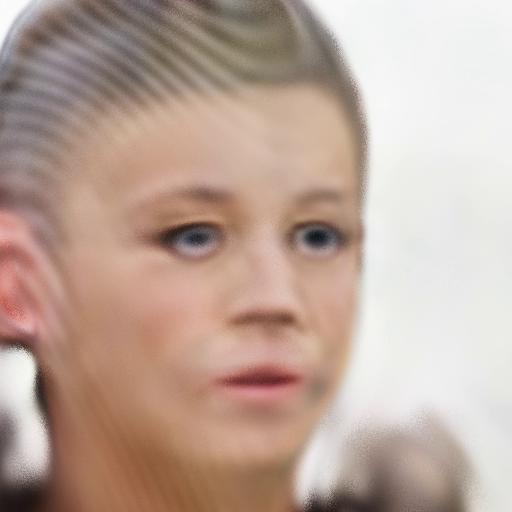} &
      \includegraphics[width=\linewidth]{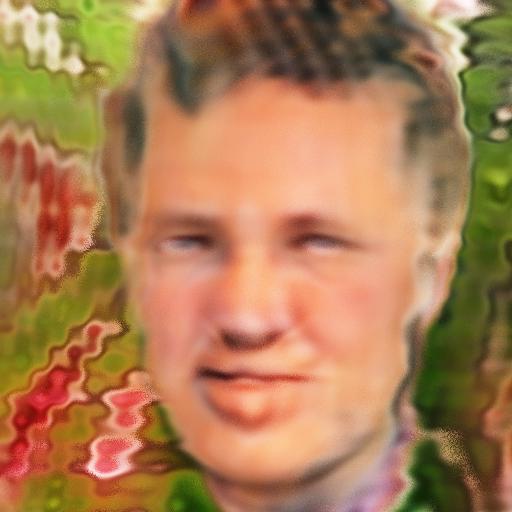} &
      \includegraphics[width=\linewidth]{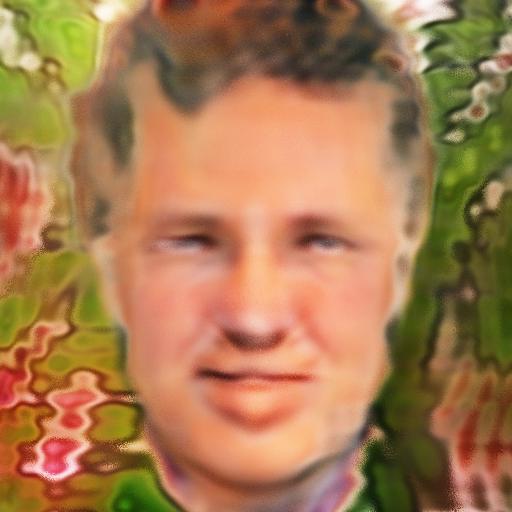} &
      \includegraphics[width=\linewidth]{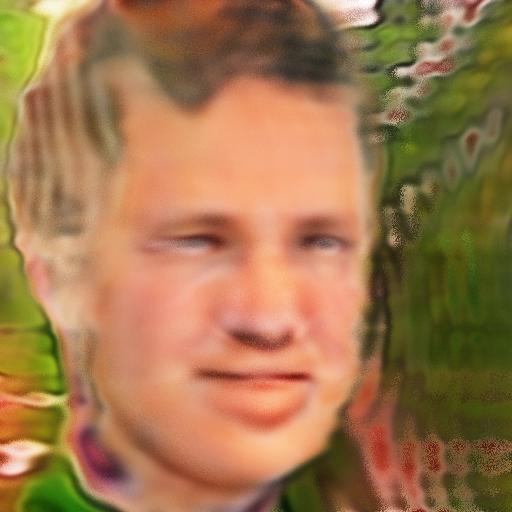}
      \\
      \includegraphics[width=\linewidth]{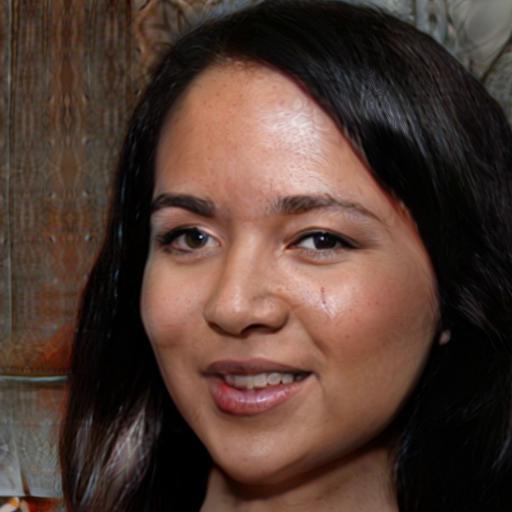}  &
      \includegraphics[width=\linewidth]{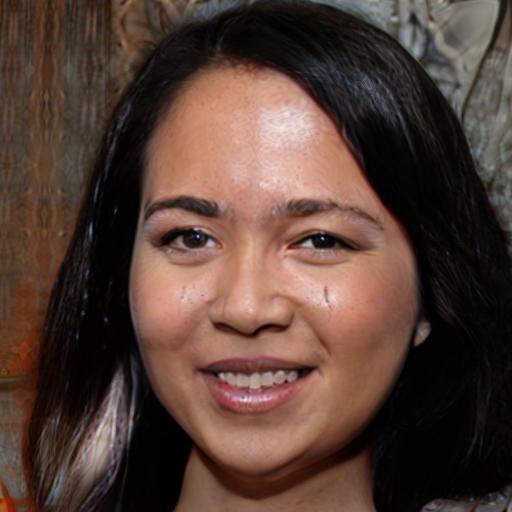}  &
      \includegraphics[width=\linewidth]{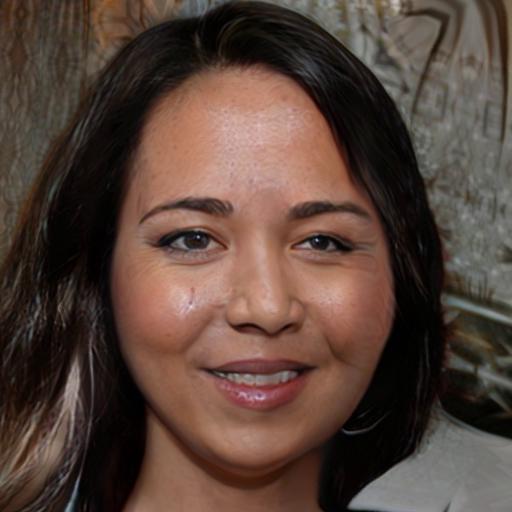}  &
      \includegraphics[width=\linewidth]{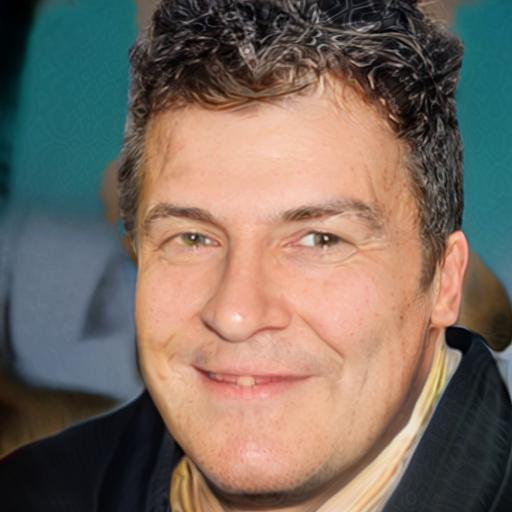}  &
      \includegraphics[width=\linewidth]{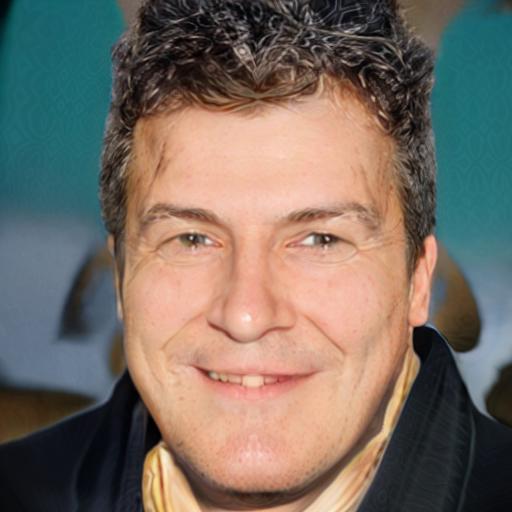}  &
      \includegraphics[width=\linewidth]{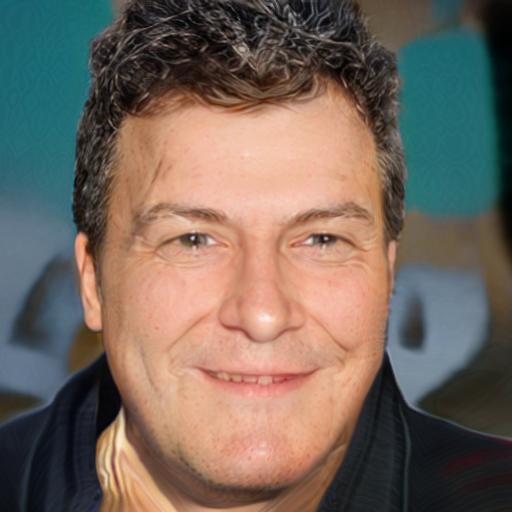}
    \end{tabular}
  }
  \vspace{-0.3cm}
  \caption{Top: Images synthesized by the NeRF network. Bottom: Corresponding images synthesized by the INR network. }
  \label{fig:nerf_inr_imgs}
  \vspace{-0.5cm}
\end{figure}

\begin{figure*}[!t]
  \footnotesize
  \centering
  \resizebox{\linewidth}{!}{%
    \renewcommand{\tabcolsep}{1pt} \renewcommand{\arraystretch}{0.9}
    \centering
    \begin{tabular}{c}

      \resizebox{\linewidth}{!}{%
        \renewcommand{\tabcolsep}{1pt} \renewcommand{\arraystretch}{0.9}
        \centering
        \begin{tabular}{cccc|cccc}
          \resizebox{!}{3cm}{
            \begin{tabular}[b]{c}
              \multicolumn{1}{c}{\includegraphics[width=\linewidth]{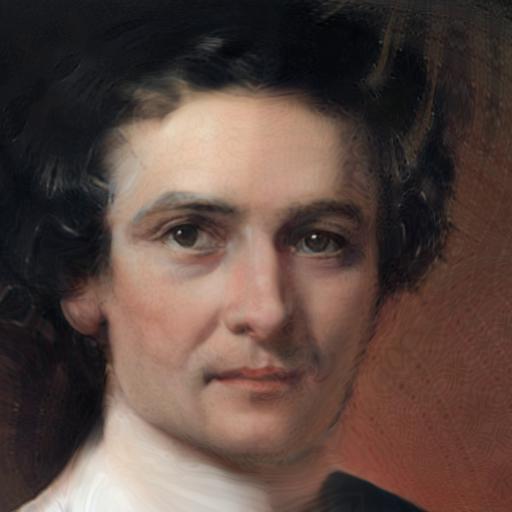}}
            \end{tabular}%
          }
                                           &
          \resizebox{!}{3cm}{%
            \begin{tabular}[b]{c}
              \multicolumn{1}{c}{\includegraphics[width=\linewidth]{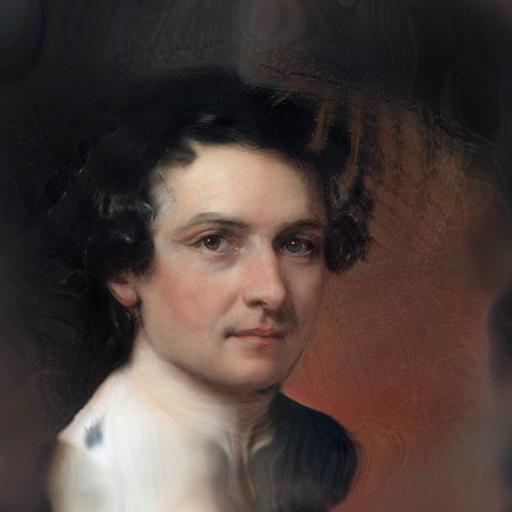}}%
              \\
              \multicolumn{1}{c}{\includegraphics[width=\linewidth]{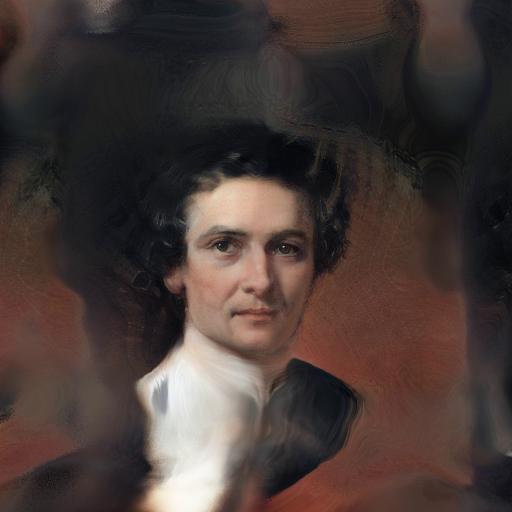}}
            \end{tabular}%
          }
                                           &
          \resizebox{!}{3cm}{
            \begin{tabular}[b]{c}
              \multicolumn{1}{c}{\includegraphics[width=\linewidth]{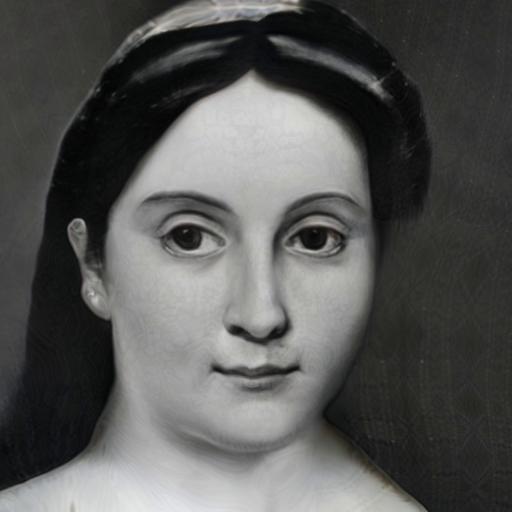}}
            \end{tabular}%
          }
                                           &
          \resizebox{!}{3cm}{%
            \begin{tabular}[b]{c}
              \multicolumn{1}{c}{\includegraphics[width=\linewidth]{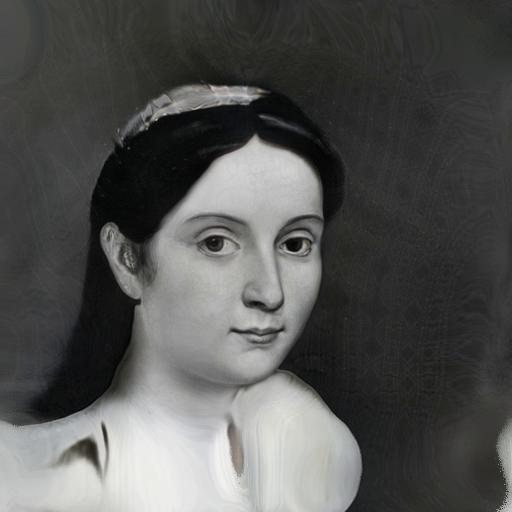}}%
              \\
              \multicolumn{1}{c}{\includegraphics[width=\linewidth]{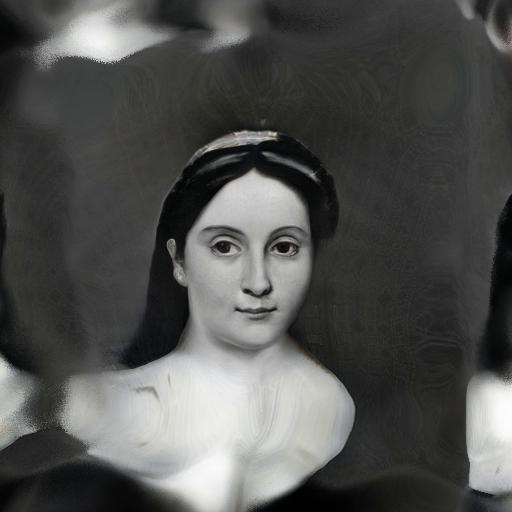}}
            \end{tabular}%
          }

          \hspace{0.01cm}                  & \hspace{0.01cm}
          \resizebox{!}{3cm}{
            \begin{tabular}[b]{c}
              \multicolumn{1}{c}{\includegraphics[width=\linewidth]{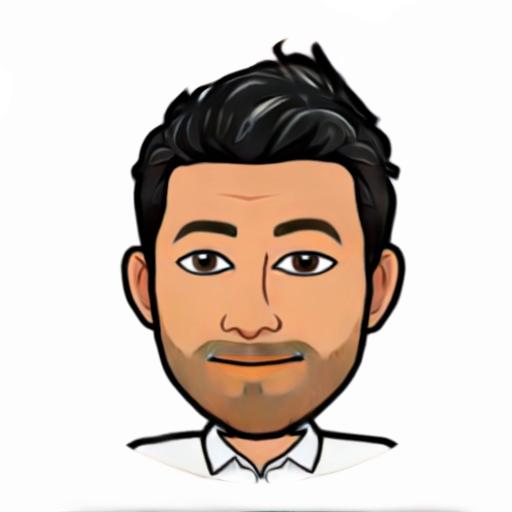}}
            \end{tabular}%
          }
                                           &
          \resizebox{!}{3cm}{%
            \begin{tabular}[b]{c}
              \multicolumn{1}{c}{\includegraphics[width=\linewidth]{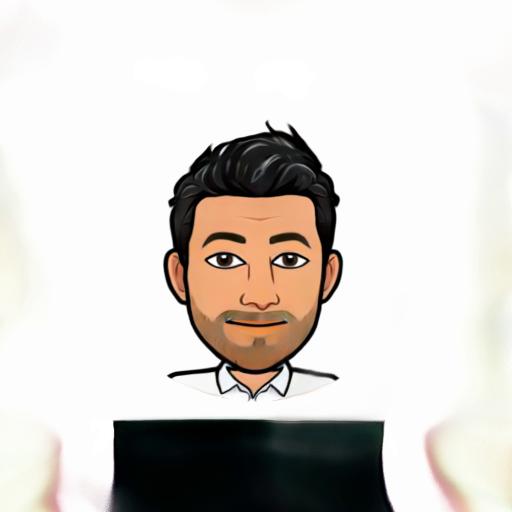}}%
              \\
              \multicolumn{1}{c}{\includegraphics[width=\linewidth]{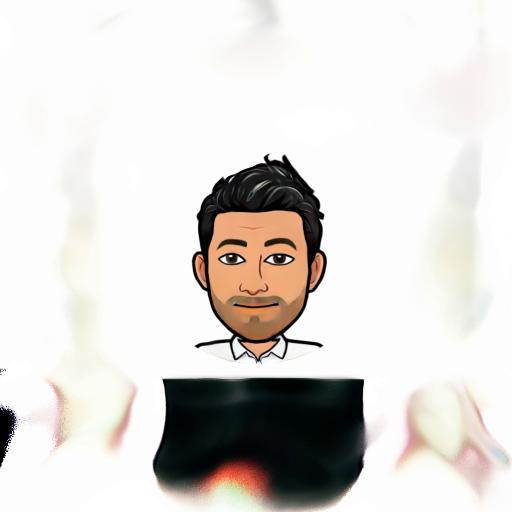}}
            \end{tabular}%
          }
                                           &
          \resizebox{!}{3cm}{
            \begin{tabular}[b]{c}
              \multicolumn{1}{c}{\includegraphics[width=\linewidth]{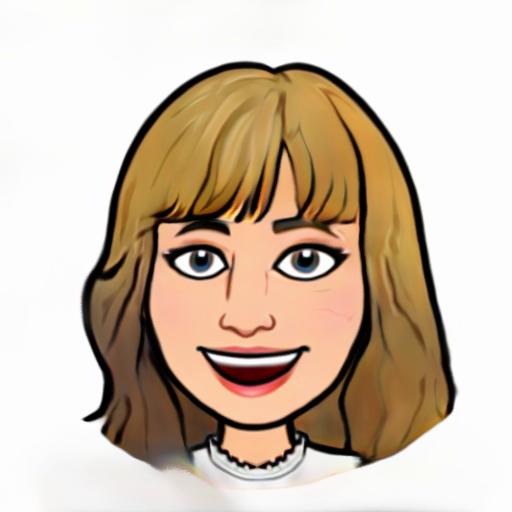}}
            \end{tabular}%
          }
                                           &
          \resizebox{!}{3cm}{%
            \begin{tabular}[b]{c}
              \multicolumn{1}{c}{\includegraphics[width=\linewidth]{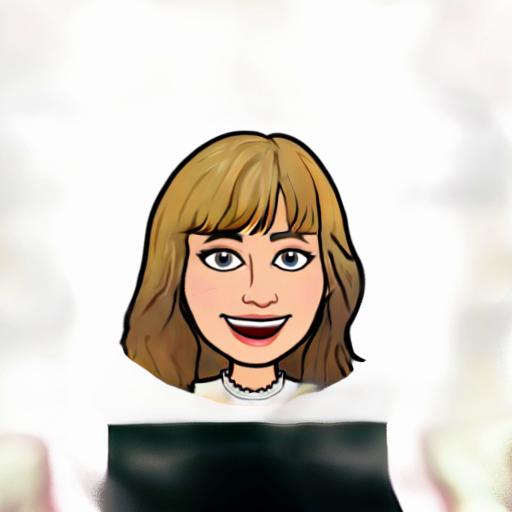}}%
              \\
              \multicolumn{1}{c}{\includegraphics[width=\linewidth]{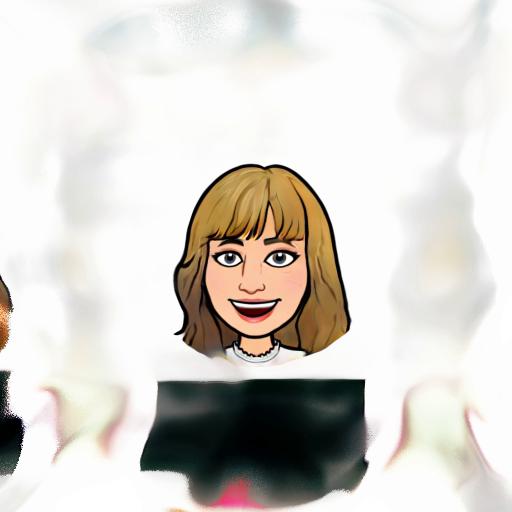}}
            \end{tabular}%
          }
          \\
          \multicolumn{4}{c}{MetFaces}     & \multicolumn{4}{c}{BitmojiFaces}
          \\
          \resizebox{!}{3cm}{
            \begin{tabular}[b]{c}
              \multicolumn{1}{c}{\includegraphics[width=\linewidth]{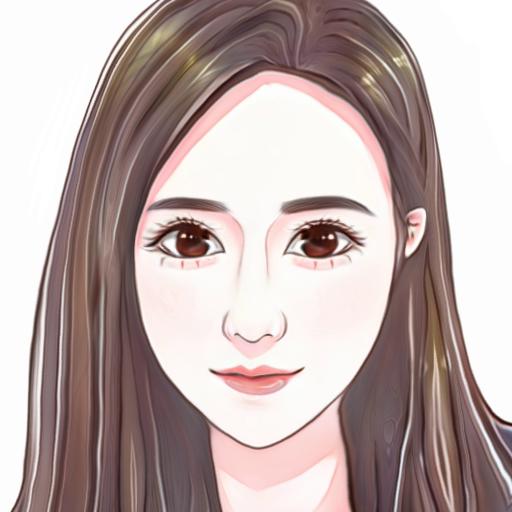}}
            \end{tabular}%
          }
                                           &
          \resizebox{!}{3cm}{%
            \begin{tabular}[b]{c}
              \multicolumn{1}{c}{\includegraphics[width=\linewidth]{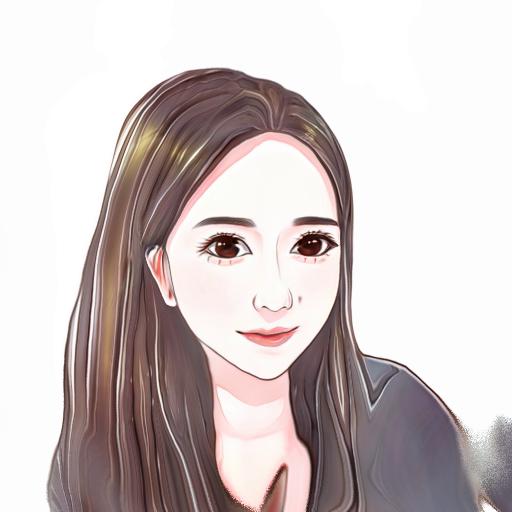}}%
              \\
              \multicolumn{1}{c}{\includegraphics[width=\linewidth]{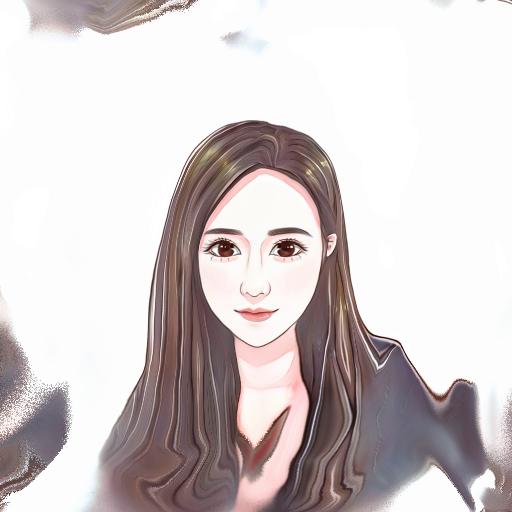}}
            \end{tabular}%
          }
                                           &
          \resizebox{!}{3cm}{
            \begin{tabular}[b]{c}
              \multicolumn{1}{c}{\includegraphics[width=\linewidth]{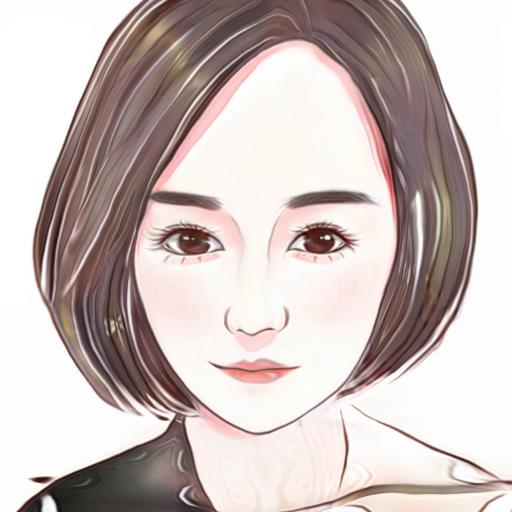}}
            \end{tabular}%
          }
                                           &
          \resizebox{!}{3cm}{%
            \begin{tabular}[b]{c}
              \multicolumn{1}{c}{\includegraphics[width=\linewidth]{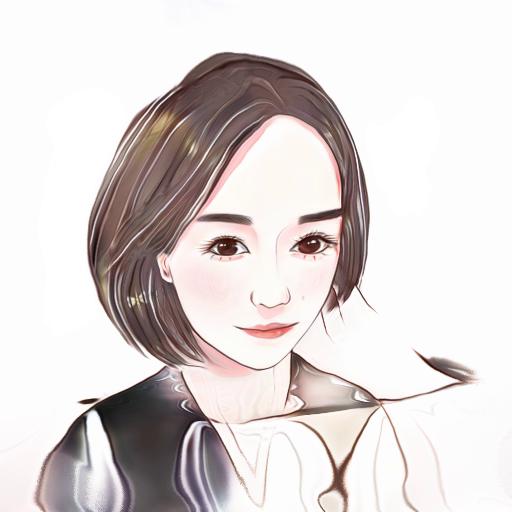}}%
              \\
              \multicolumn{1}{c}{\includegraphics[width=\linewidth]{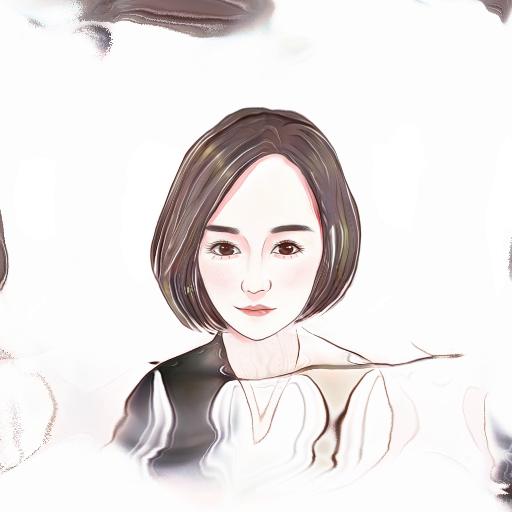}}
            \end{tabular}%
          }

          \hspace{0.01cm}                  & \hspace{0.01cm}
          \resizebox{!}{3cm}{
            \begin{tabular}[b]{c}
              \multicolumn{1}{c}{\includegraphics[width=\linewidth]{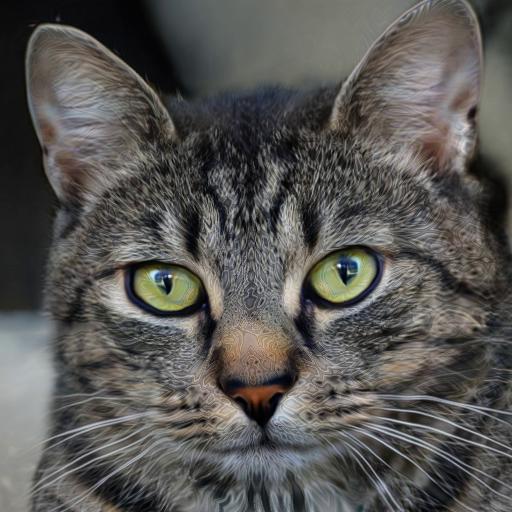}}
            \end{tabular}%
          }
                                           &
          \resizebox{!}{3cm}{%
            \begin{tabular}[b]{c}
              \multicolumn{1}{c}{\includegraphics[width=\linewidth]{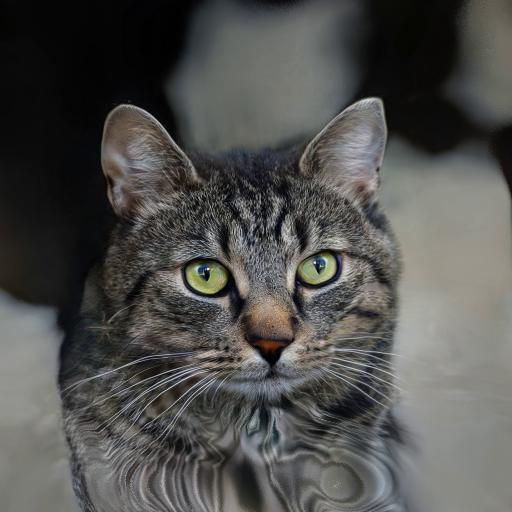}}%
              \\
              \multicolumn{1}{c}{\includegraphics[width=\linewidth]{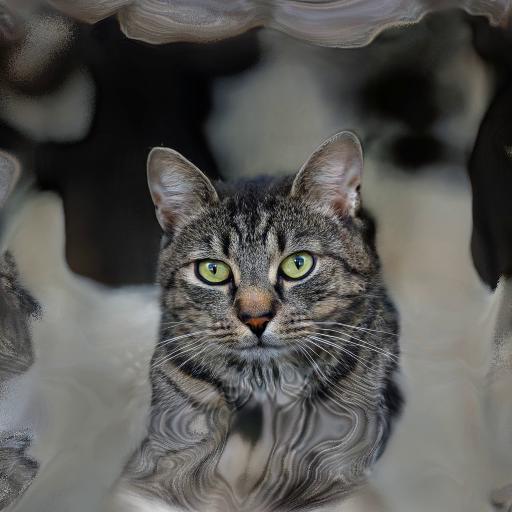}}
            \end{tabular}%
          }
                                           &
          \resizebox{!}{3cm}{
            \begin{tabular}[b]{c}
              \multicolumn{1}{c}{\includegraphics[width=\linewidth]{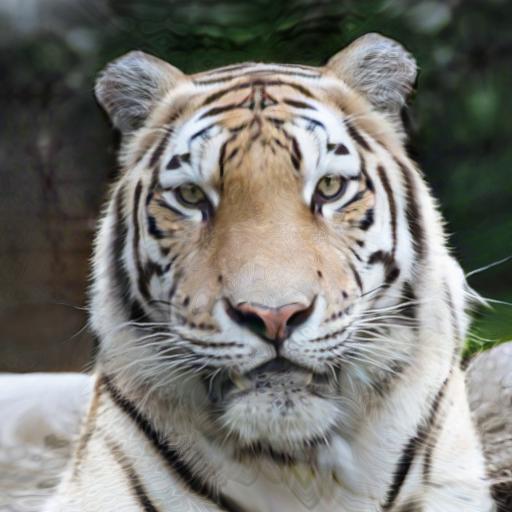}}
            \end{tabular}%
          }
                                           &
          \resizebox{!}{3cm}{%
            \begin{tabular}[b]{c}
              \multicolumn{1}{c}{\includegraphics[width=\linewidth]{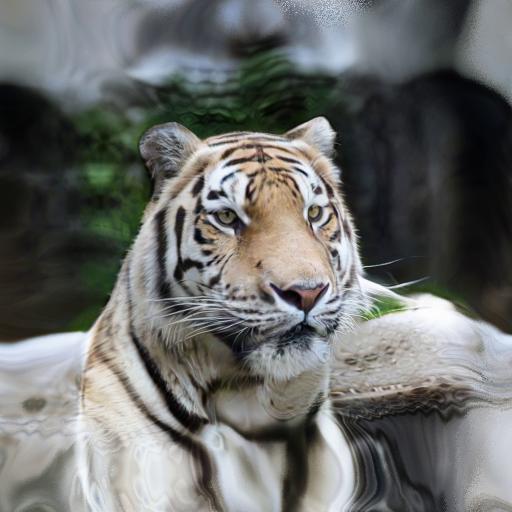}}%
              \\
              \multicolumn{1}{c}{\includegraphics[width=\linewidth]{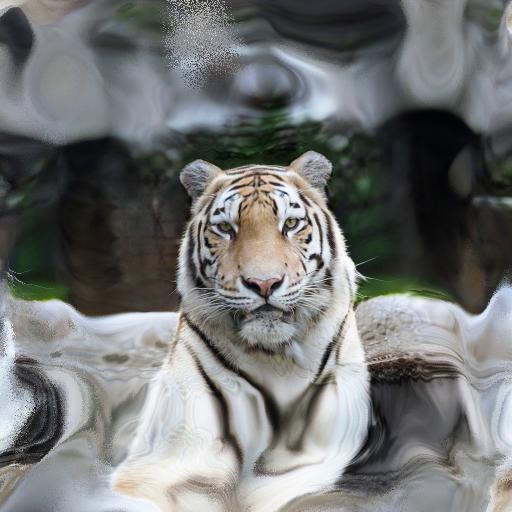}}
            \end{tabular}%
          }
          \\
          \multicolumn{4}{c}{CartoonFaces} & \multicolumn{4}{c}{AFHQ}
          \\
          \multicolumn{8}{c}{\resizebox{!}{0.25cm}{(a) Images synthesized by transferred models.}}
        \end{tabular}
      }
      \\
      \midrule
      \resizebox{\linewidth}{!}{%
        \renewcommand{\tabcolsep}{3pt} \renewcommand{\arraystretch}{3}
        \centering
        \begin{tabular}{c|cccccccc}
          \includegraphics[width=\linewidth]{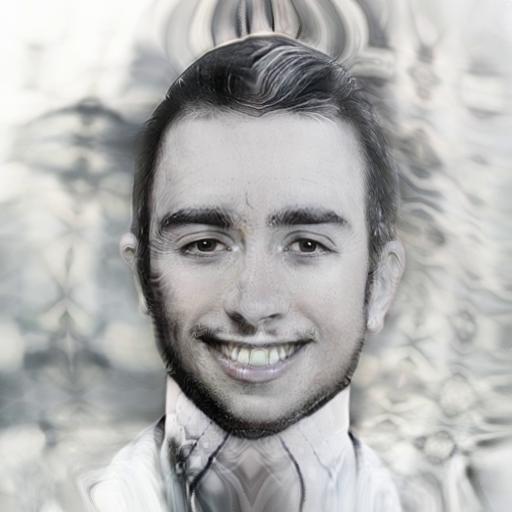}
          \includegraphics[width=\linewidth]{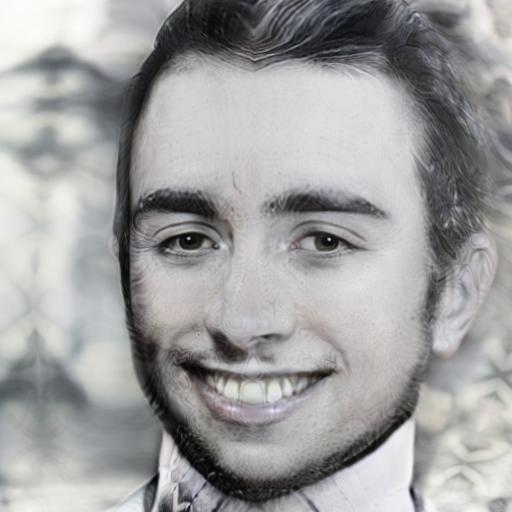}  \hspace{0.4cm}  & \hspace{0.4cm}
          \includegraphics[width=\linewidth]{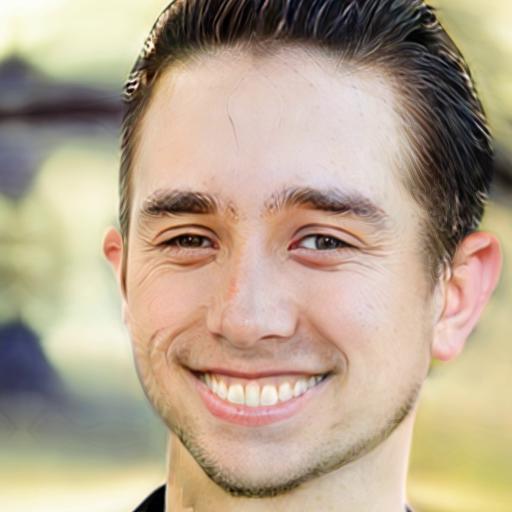}                 &
          \includegraphics[width=\linewidth]{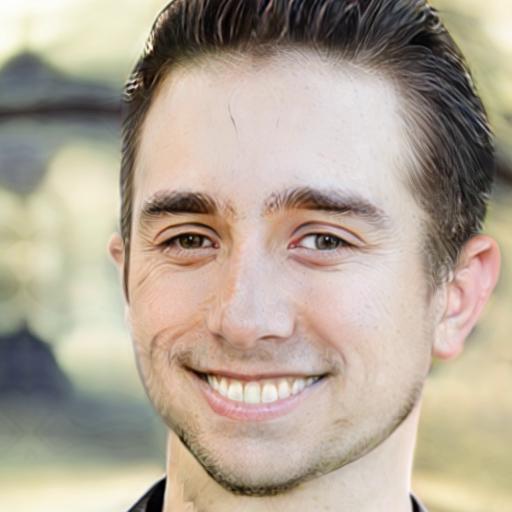}                 &
          \includegraphics[width=\linewidth]{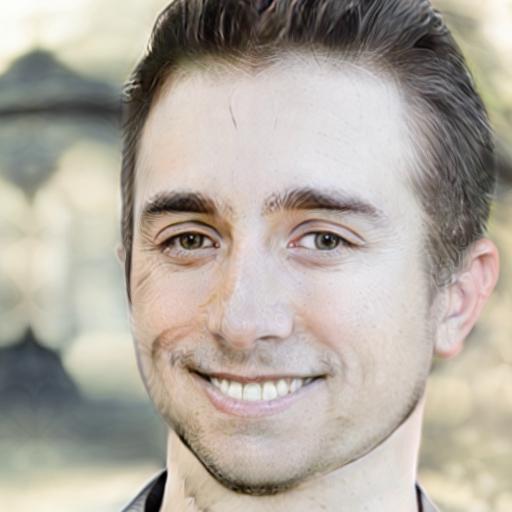}                 &
          \includegraphics[width=\linewidth]{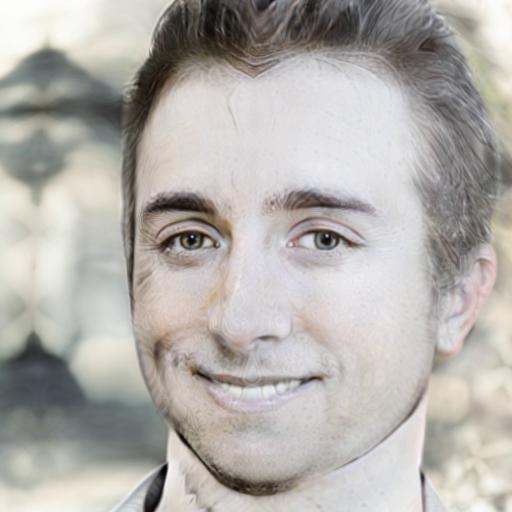}                 &
          \includegraphics[width=\linewidth]{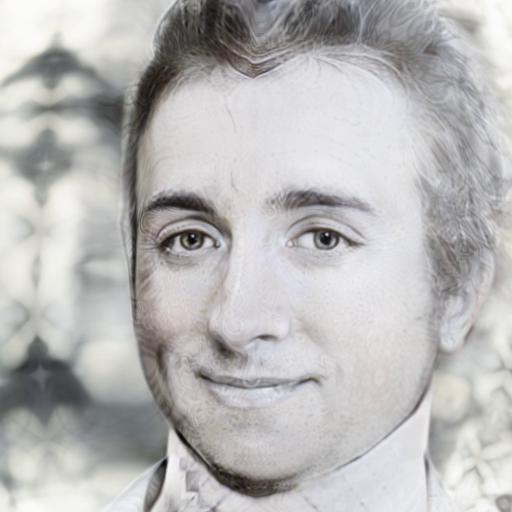}                 &
          \includegraphics[width=\linewidth]{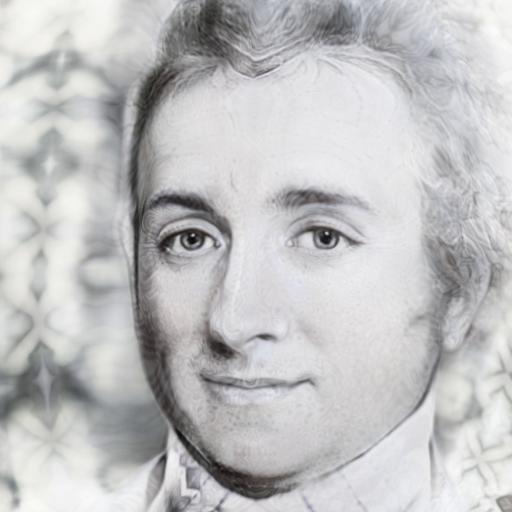}                 &
          \includegraphics[width=\linewidth]{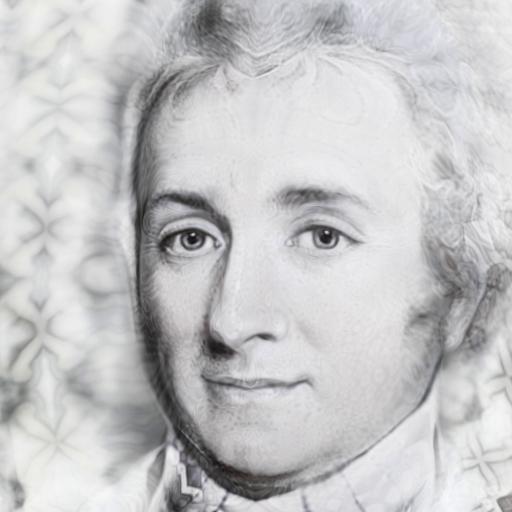}                 &
          \includegraphics[width=\linewidth]{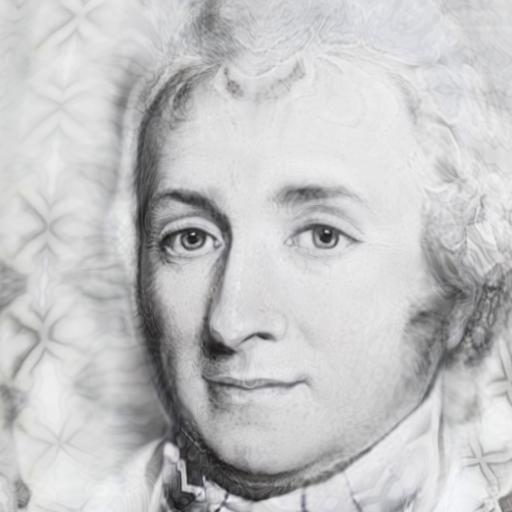}
          \\ 
          \includegraphics[width=\linewidth]{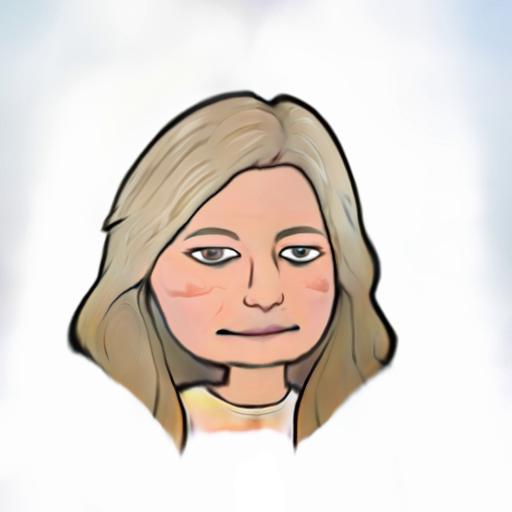}
          \includegraphics[width=\linewidth]{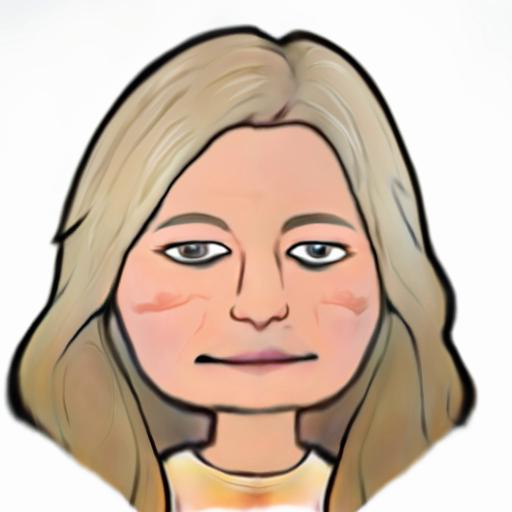}  \hspace{0.4cm} & \hspace{0.4cm}
          \includegraphics[width=\linewidth]{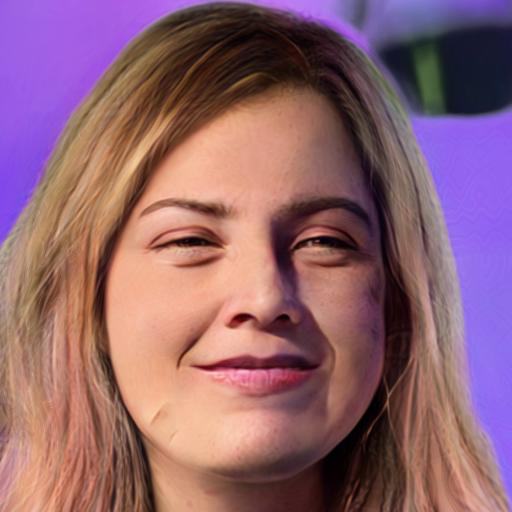}                &
          \includegraphics[width=\linewidth]{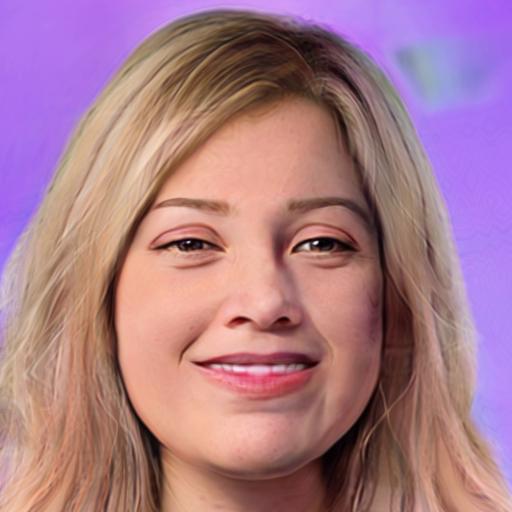}                &
          \includegraphics[width=\linewidth]{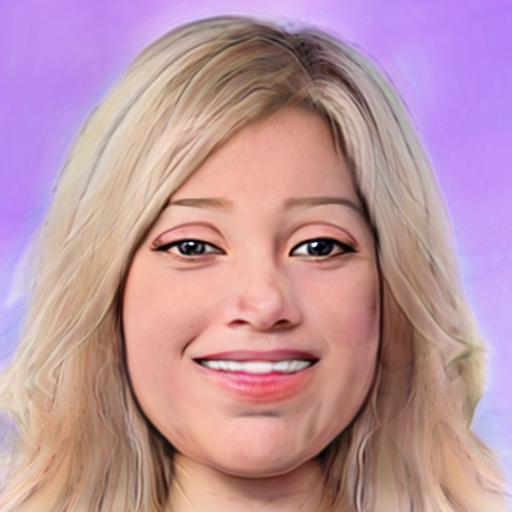}                &
          \includegraphics[width=\linewidth]{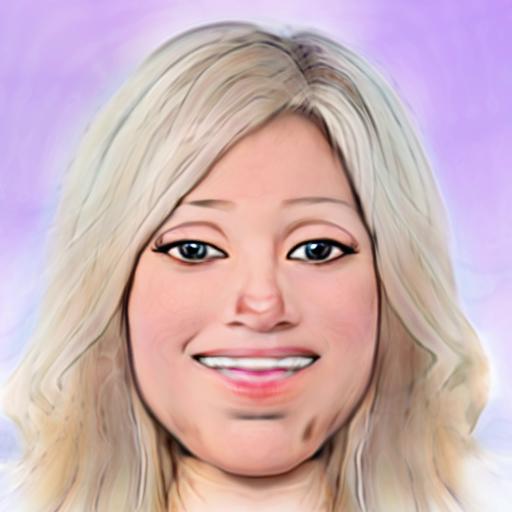}                &
          \includegraphics[width=\linewidth]{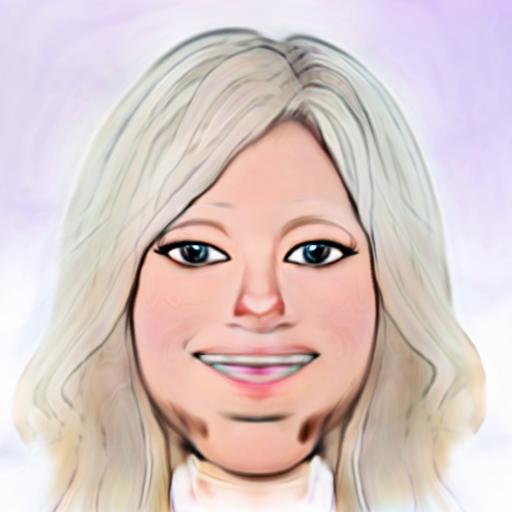}                &
          \includegraphics[width=\linewidth]{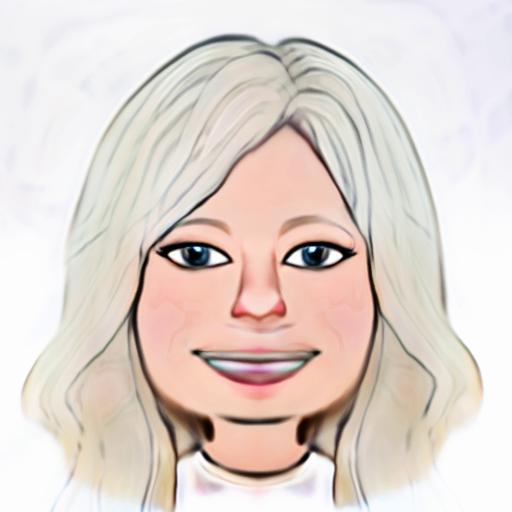}                &
          \includegraphics[width=\linewidth]{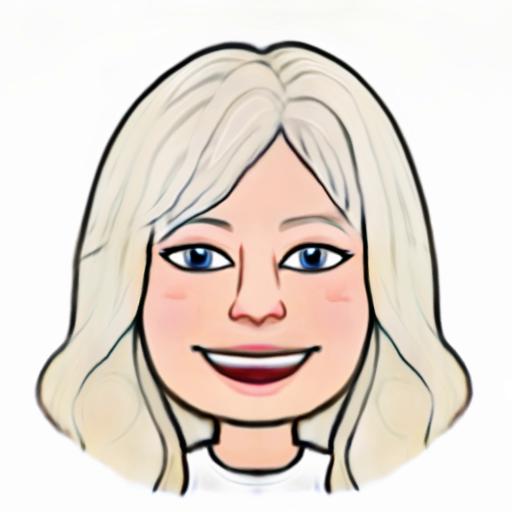}                &
          \includegraphics[width=\linewidth]{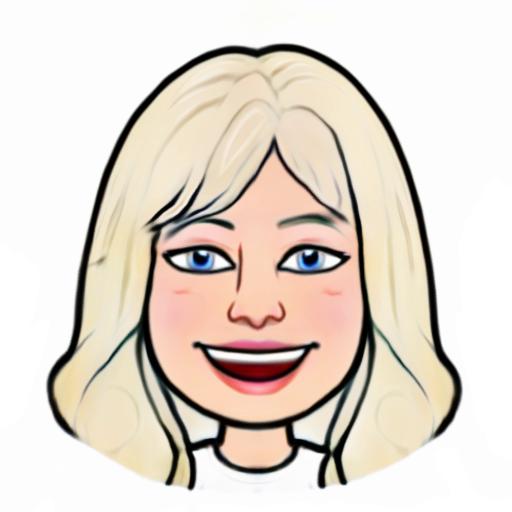}
          \\ 
          \includegraphics[width=\linewidth]{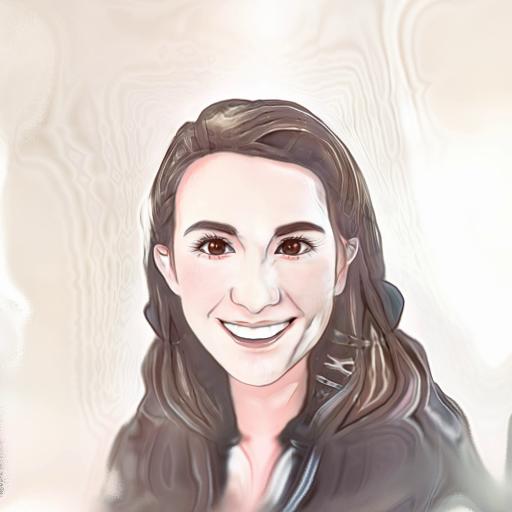}
          \includegraphics[width=\linewidth]{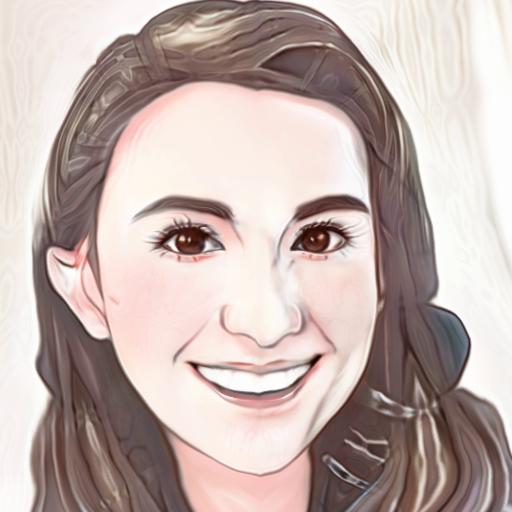}  \hspace{0.4cm}  & \hspace{0.4cm}
          \includegraphics[width=\linewidth]{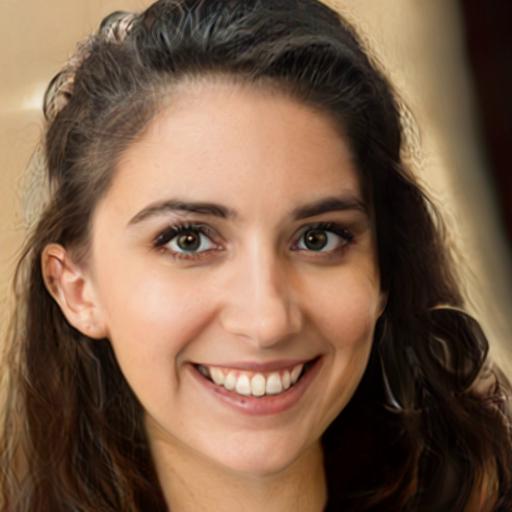}                 &
          \includegraphics[width=\linewidth]{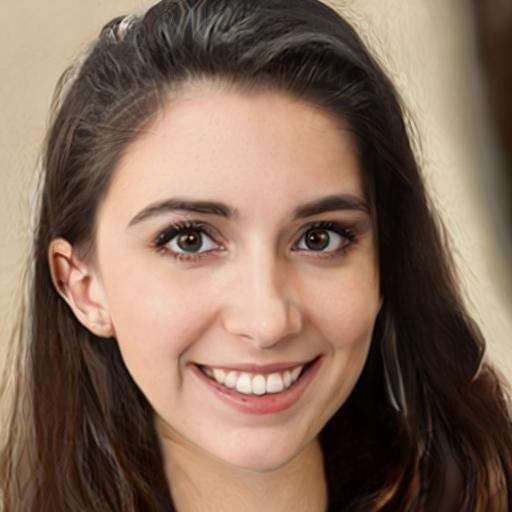}                 &
          \includegraphics[width=\linewidth]{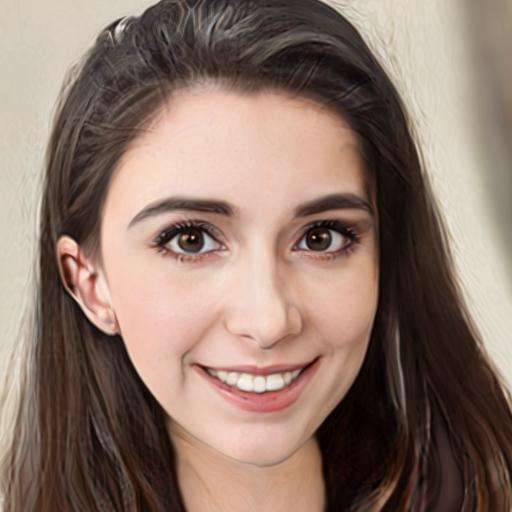}                 &
          \includegraphics[width=\linewidth]{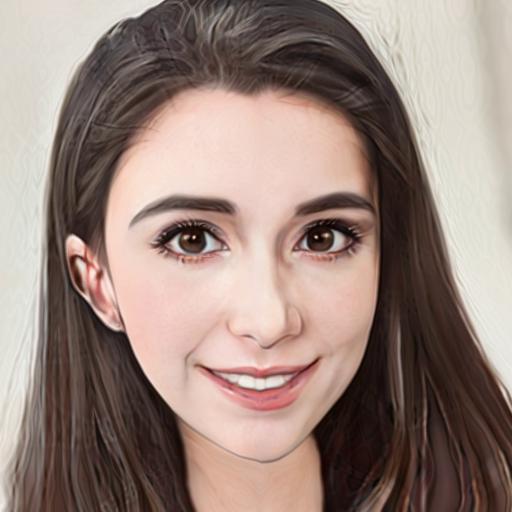}                 &
          \includegraphics[width=\linewidth]{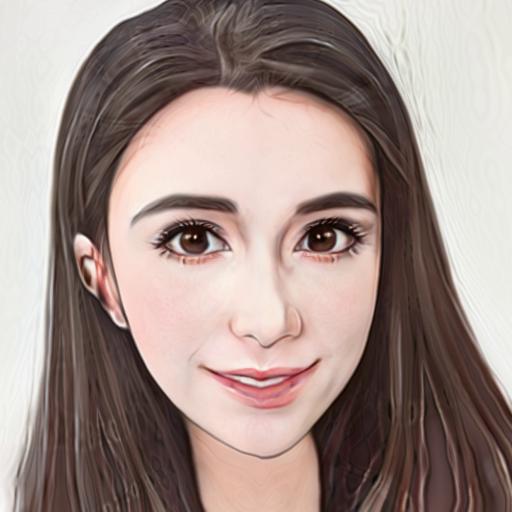}                 &
          \includegraphics[width=\linewidth]{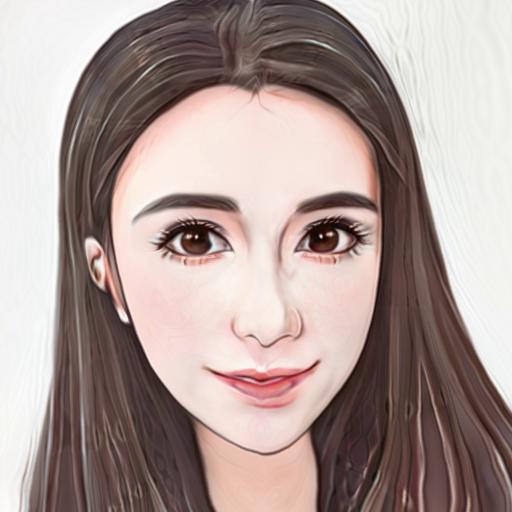}                 &
          \includegraphics[width=\linewidth]{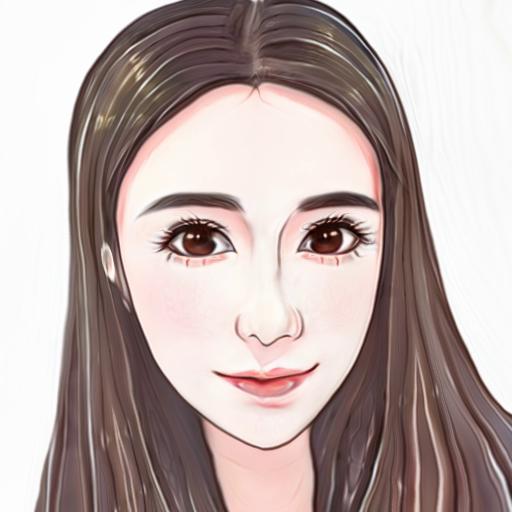}                 &
          \includegraphics[width=\linewidth]{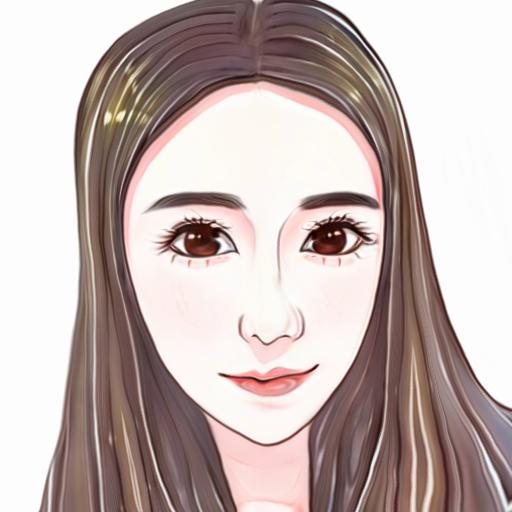}
          \\ 
          \includegraphics[width=\linewidth]{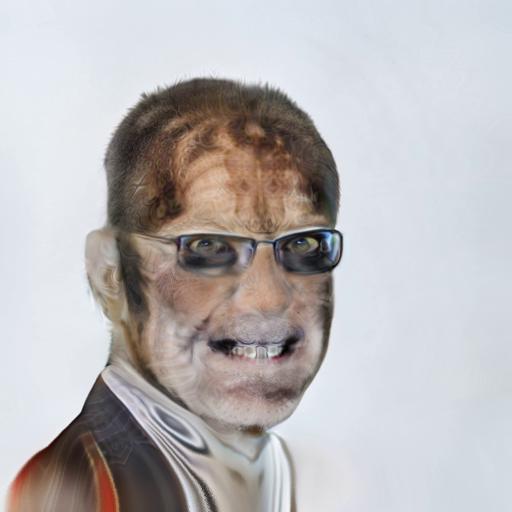}
          \includegraphics[width=\linewidth]{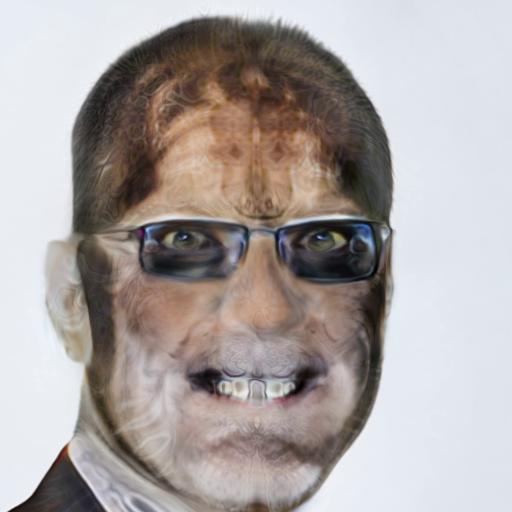}  \hspace{0.4cm}  & \hspace{0.4cm}
          \includegraphics[width=\linewidth]{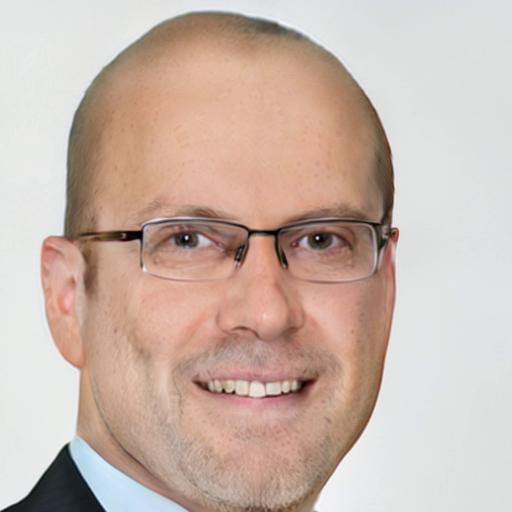}                 &
          \includegraphics[width=\linewidth]{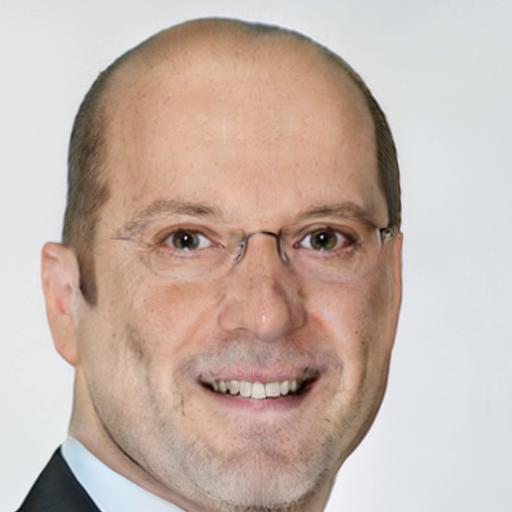}                 &
          \includegraphics[width=\linewidth]{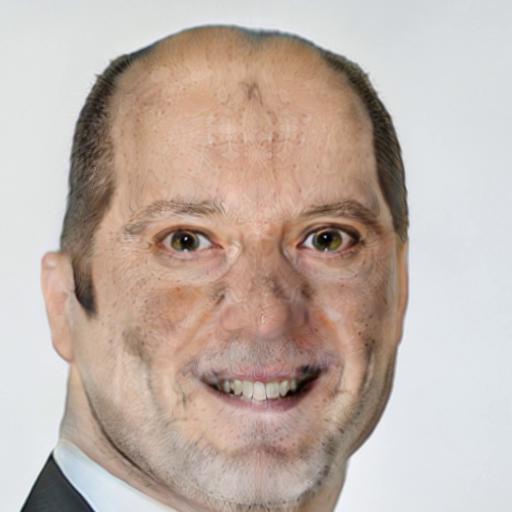}                 &
          \includegraphics[width=\linewidth]{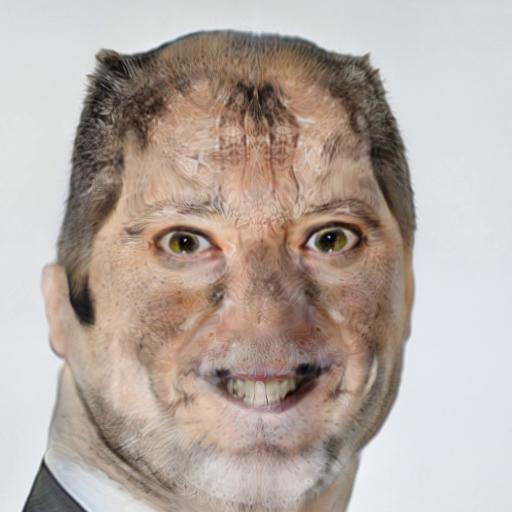}                 &
          \includegraphics[width=\linewidth]{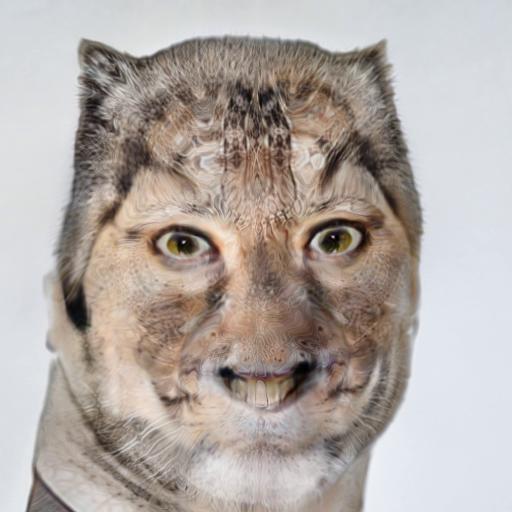}                 &
          \includegraphics[width=\linewidth]{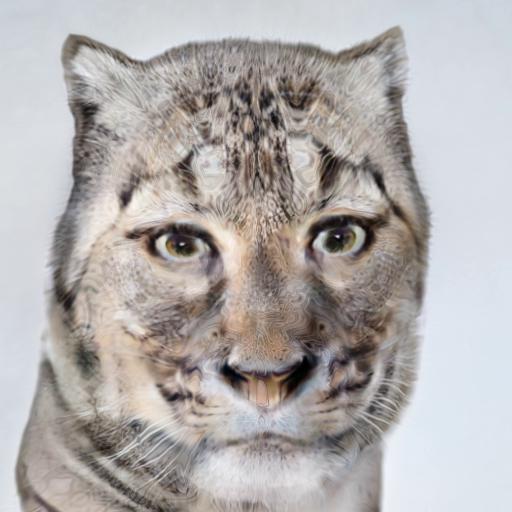}                 &
          \includegraphics[width=\linewidth]{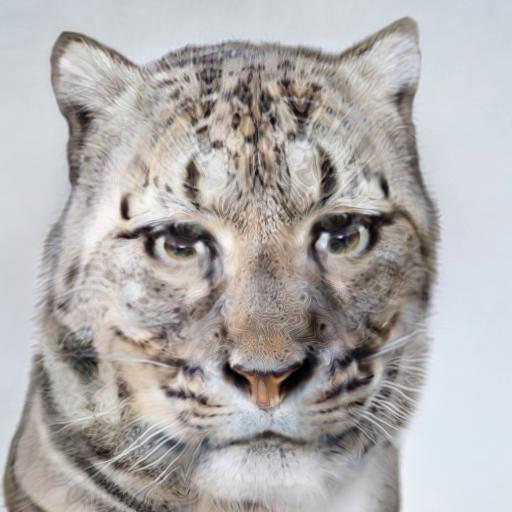}                 &
          \includegraphics[width=\linewidth]{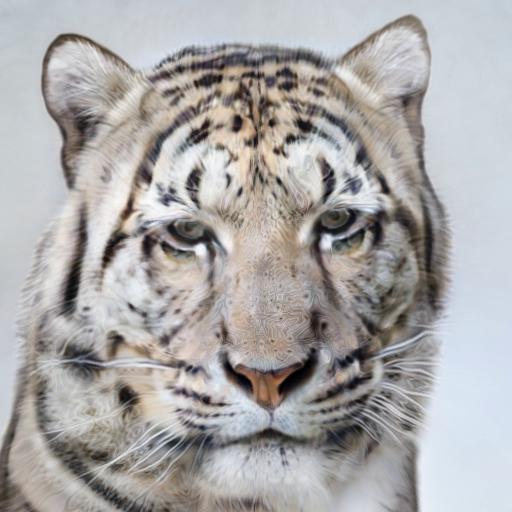}
          \\
          \resizebox{!}{2.0cm}{(b) Stylization (multi-views)}                                                        & \resizebox{!}{1.5cm}{Content} & \multicolumn{6}{c}{\resizebox{!}{2.2cm}{(c) Images synthesized by interpolated models (linear interpolation).}} & \resizebox{!}{1.5cm}{Style}
        \end{tabular}
      }
    \end{tabular}
  }
  \vspace{-0.3cm}
  \caption{(a): Fine-tuning the base model trained on FFHQ to generate images in other domains (please refer to~\cref{sec:finetune} for details). (b) and (c): Interpolating the base model and the transferred model to generate stylized images (please refer to~\cref{sec:interpolation} for details). CIPS-3D enables us to manipulate the pose of the generated faces explicitly. }
  \label{fig:finetune_and_interpolation}
  \vspace{-0.5cm}
\end{figure*}

\subsection{Transfer Learning}
\label{sec:finetune}

In~\cref{fig:nerf_inr_imgs}, we show the output images of the NeRF network and the corresponding output images of the INR network. The INR output images share the same pose as the NeRF images but own much richer textures. This indicates that the NeRF network is responsible for the posture and the INR network is responsible for the texture. Inspired by the FreezeG~\cite{FreezeG} which freezes the early layers of the generator, we freeze the NeRF network of our generator trained on FFHQ and only fine-tune the INR network for transfer learning settings. Freezing NeRF is critical for transfer learning because there are too few images in the target domain. It is almost impossible to learn 3D shapes from so few images. However, by reusing the weights of NeRF, we only need to update the 2D INR network to render textures of other domains.

We verified the efficacy of the method on four datasets, including MetFaces~\cite{karras2020Training}, BitmojiFaces~\cite{BitmojiFaces}, CartoonFaces~\cite{CartoonFaces}, and even the animal dataset, AFHQ~\cite{choi2020StarGAN}. As shown in~\cref{fig:finetune_and_interpolation}a, the transferred models generate images of different domains. Moreover, we can easily control the pose of the generated faces by explicitly manipulating the NeRF network.


\subsection{3D-aware Face Stylization}
\label{sec:interpolation}

We consider interpolating the original base model trained on FFHQ and the transferred models (fine-tuned on target datasets) to create new models to generate images in novel domains. Note that the weights of the NeRF network of the base mode and the transferred models are completely equal. We test two types of interpolating methods: (i) linearly interpolating all INR layers, and (ii) replacing the higher layers of the INR network of the base model with the corresponding layers of the transferred model~\cite{pinkney2020Resolution,huang2020Unsupervised}. The former produces images between two domains, as shown in~\cref{fig:finetune_and_interpolation}c. The images smoothly fade from one domain to another. The latter generates images that combine the structural characteristics of the content images and the style characteristics of the style images (see~\cref{fig:finetune_and_interpolation}b). Besides, CIPS-3D allows one to manipulate the pose of all the stylized faces explicitly.


\vspace{-0.1cm}
\section{Limitation and Future Work}
\vspace{-0.1cm}
CIPS-3D works in a noise-to-image manner. Thus current stylization is limited to randomly generated images. To edit a real image, we need to project the image to the latent space of the generator. For this purpose, we need to study 3D-aware GAN inversion, and we leave this to future work.


\vspace{-0.1cm}
\section{Conclusion}
\vspace{-0.1cm}

This paper presents a style-based 3D-aware generator that synthesizes pixel values independently without any spatial convolution or up-sampling operation. We find that the symmetry of the input coordinates leads to the problem of mirror symmetry and propose to utilize an auxiliary discriminator to solve this problem. We look forward to applying the proposed generator to more interesting applications such as 3D-aware GAN inversion and image-to-image translation.

\clearpage
{\small
  \bibliographystyle{ieee_fullname}
  \bibliography{egbib}
}

\clearpage

\end{document}


\twocolumn[{%
\centering
\Large\textbf{``CIPS-3D: A 3D-Aware Generator of GANs Based on \\Conditionally-Independent Pixel Synthesis'' Supplementary Material}
\\ [1.5em]
}]


\appendix
\addcontentsline{toc}{section}{Appendices}


